\documentclass[]{interact}

\usepackage{epstopdf}
\usepackage[caption=false]{subfig}

\usepackage[numbers,sort&compress]{natbib}
\bibpunct[, ]{[}{]}{,}{n}{,}{,}
\renewcommand\bibfont{\fontsize{10}{12}\selectfont}
\makeatletter
\def\NAT@def@citea{\def\@citea{\NAT@separator}}
\makeatother

\theoremstyle{plain}

\theoremstyle{definition}

\theoremstyle{remark}

\usepackage{url}
\usepackage{gensymb}

\newcommand*{\vertbar}{\rule[-1ex]{0.5pt}{2.5ex}}

\newcommand*{\horzbar}{\rule[.5ex]{2.5ex}{0.5pt}}

\begin{document}

\articletype{FULL ARTICLE}

\title{RUR53: an Unmanned Ground Vehicle for Navigation, Recognition and Manipulation}

\author{
\name{Nicola Castaman\thanks{CONTACT Nicola Castaman Email: nicola.castaman@dei.unipd.it}, Elisa Tosello, Morris Antonello, Nicola Bagarello, Silvia Gandin, Marco Carraro, Matteo Munaro, Roberto Bortoletto, Stefano Ghidoni, Emanuele Menegatti and Enrico Pagello}
\affil{Intelligent Autonomous Systems Lab (IAS-Lab), Department of Information Engineering, University of Padova, Via Gradenigo 6/B, 35131 Padova, Italy}
}

\maketitle

\begin{abstract}
This paper proposes RUR53: an Unmanned Ground Vehicle able to autonomously navigate through, identify, and reach areas of interest; and there recognize, localize, and manipulate work tools to perform complex manipulation tasks. The proposed contribution includes a modular software architecture where each module solves specific sub-tasks and that can be easily enlarged to satisfy new requirements.
Included indoor and outdoor tests demonstrate the capability of the proposed system to autonomously detect a target object (a panel) and precisely dock in front of it while avoiding obstacles.
They show it can autonomously recognize and manipulate target work tools (i.e., wrenches and valve stems) to accomplish complex tasks (i.e., use a wrench to rotate a valve stem).
A specific case study is described where the proposed modular architecture lets easy switch to a semi-teleoperated mode.
The paper exhaustively describes description of both the hardware and software setup of RUR53, its performance when tests at the 2017 Mohamed Bin Zayed International Robotics Challenge, and the lessons we learned when participating at this competition, where we ranked third in the Gran Challenge in collaboration with the Czech Technical University in Prague, the University of Pennsylvania, and the University of Lincoln (UK).
\end{abstract}


\section{Introduction}\label{introduction}
This paper presents RUR53: an Unmanned Ground Vehicle (UGV) able to autonomously navigates in large areas, recognizes tools, and manipulates them in order to perform specific tasks.
Strengths of our system include:
\begin{itemize}
    \item \textit{Modularity}: 
    RUR53 software architecture is modular and each module fulfills a specific task. Such a structure favors stability of the system and facilitates code re-usability.
    \item \textit{Generality}: 
    RUR53 can freely navigate within any indoor and outdoor environment. The perception routine can be easily trained and used to identify any object; moreover, the 3-finger gripper lets the robot grasp objects of any geometry and dimension. 
    \item \textit{Reliability}: RUR53 is designed to be reliable in outdoor scenarios.
    It is equipped with two outdoor 2D Light Detection And Ranging (LiDAR) sensors, covering all robot surroundings. A stereo vision system, providing both 2D and 3D data of detected objects makes the robot robust to perception errors and light variations. The software architecture is developed to handle task failure.
\end{itemize}

RUR53 has been developed in the context of the 2017 Mohamed Bin Zayed International Robotics Challenge\footnote{See \url{http://www.mbzirc.com/}} (MBZIRC 2017), an international robotics competition that  took place in Abu Dhabi in March 2017 and that consisted of three challenges and a triathlon type Grand Challenge, which combined the first three. 
Our team competed in Challenge 2 (and its corresponding part in the Grand Challenge), where a UGV had to locate and reach a panel in a outdoor $60\times60$~m arena, select and grasp an assigned wrench on the panel, and use it to turn a valve stem on the panel itself. 
To this aim, the software architecture of RUR53 lets tha autonomous navigation and exploration of an outdoor arena, the detection of a panel, and the dock of the robot in front of it. Moreover, it lets the recognition of wrenches and valves despite the noise introduced by the reflection of the sun on the metal surface of these tools and even if located at different distances from the panel. Finally, it lets the manipulation of these tools.
The modularity of the proposed system allowed us to successfully and quickly face the drawbacks encountered when testing the system in Abu Dhabi. Indeed, while in the lab the proposed navigation routine was able to autonomously detect and reach the panel, in Abu Dhabi the arena configuration affected this process, as discussed in the Lessons Learned (Section~\ref{lessons_learned}). Nevertheless, we were able to quickly implement a semi-teleoperated routine that let us lead the robot to the panel.
These implementation choices let us rank third in The Grand Challenge in collaboration with the Czech Technical University in Prague (Czech Republic), the University of Pennsylvania (USA), and the University of Lincoln (UK).

The paper is organized as follows. Section~\ref{related_work} gives an overview of existing robotic solutions in terms of similar tasks. Sections~\ref{hardware} and~\ref{software} detail the hardware and software architecture of RUR53. From Section~\ref{navigation} to~\ref{teleoperation}, every block of this software architecture is detailed.
Section~\ref{results_discussion} summarizes performed experiments and results obtained both in the lab and during the competition. Section~\ref{lessons_learned} details our MBZIRC experience and lessons learned, while Section~\ref{conclusion} contains conclusions and future works.

\section{Related Work}\label{related_work}

Robots are increasingly expected to autonomously face challenging situations, characterized by uncontrolled and not easily reproducible conditions. In this context, robotics competitions are spreading across the robotics community, as they propose complex experimental setups solvable through a combination of state of the art algorithms and innovative ideas~\cite{dias2016robot}. 
Autonomous navigation and manipulation are among these challenges and MBZIRC 2017 aimed to give its contribution in facing them. Focusing on MBZIRC 2017 Challenge 2, and its correspondence in the Grand Challenge, this Section compares our implementation choices to the state of the art.

The arena is not completely known; thus, RUR53 has to navigate within it
while avoiding collisions~\cite{castaman2016sampling} and accomplishing Simultaneous Localization and Mapping (SLAM) \cite{dissanayake}, all taking into consideration that there exists a mutual dependency between the map representation and the robot pose estimation: for localization, the robot needs a consistent map and, for acquiring the map, the robot requires a good estimate of its location.
Such a dependency requires searching for a solution in a high-dimensional space~\cite{grisetti2005improving}.
Focusing on map representation, the occupancy grid approach is commonly adopted~\cite{Moravec:1988:SFC:46184.46187}: it is computationally expensive but it represents the environment in detail. Focusing on the robot pose estimation, Extended Kalman Filters (EKFs) estimate a fully correlated posterior over landmark maps and robot poses. However, if no assumption is made on both the robot motion model and the sensor noise, and if no landmark is assumed to be uniquely identifiable, then the filter will diverge \cite{Frese01simultaneouslocalization}.
In~\cite{Murphy_bayesianmap},  Murphy faced this limitation by employing the Rao-Blackwellized Particle Filters (RBPF) and using each particle of RBPF to represent a possible robot trajectory and a map. \cite{fox3efficient} improved this approach by reducing the number of required particles, thus guaranteeing the capability of dealing with large environments.
Because of the dimension of the MBZIRC arena, our approach is based on \cite{fox3efficient} and an occupancy grid is adopted to represent the environment. Such a combination allows to solve SLAM for laser range data through a highly efficient RBPF \cite{grisetti2007improved,grisetti2005improving}: instead of using a fixed proposal distribution, an improved proposal distribution is computed on the fly on a per particle base. This allows to directly use most of the information obtained from the sensors, while generating the particles. In this way, particles are more effective drawn and the number of effective particles can be used to decide whether a re-sampling is necessary.

As the Challenge is outdoor (outdoor light varies) and the panal is black (black surfaces do not reflect the visible light), exploiting vision data is not enough to detect it: RUR53 exploits the laser range data.
Finally, RUR53 has to detect, recognize, and localize a set of wrenches and a valve stem, which are metallic and cause strong reflections. Among existing methods for objects detection, shape-based approaches~\cite{zhu2008contour} demonstrate strong performance when the object shape is accurately known. Hybrid approaches combining together a shape-based method and an object detector~\cite{ferrari2010images} are more generic and accurate but they are often based on iterative algorithms and are computationally expensive. We decided to adopt a Cascade of Boosted Classifier working with Local Binary Patterns (LBP) features~\cite{Felzenszwalb2010Object}, which reduces the detection time and enhances robot reactivity. 
Haar features~\cite{Viole2001Rapid,lienhart2002extended} are another common choice. Both approaches represent a good compromise between accuracy and computational workload. LBP has been selected because both the training and the detection procedures are faster then Haar (typically in the order of seconds or minutes).

\section{Hardware Architecture}\label{hardware}

\begin{figure}
  \centering
  \includegraphics[width = 0.75\linewidth]{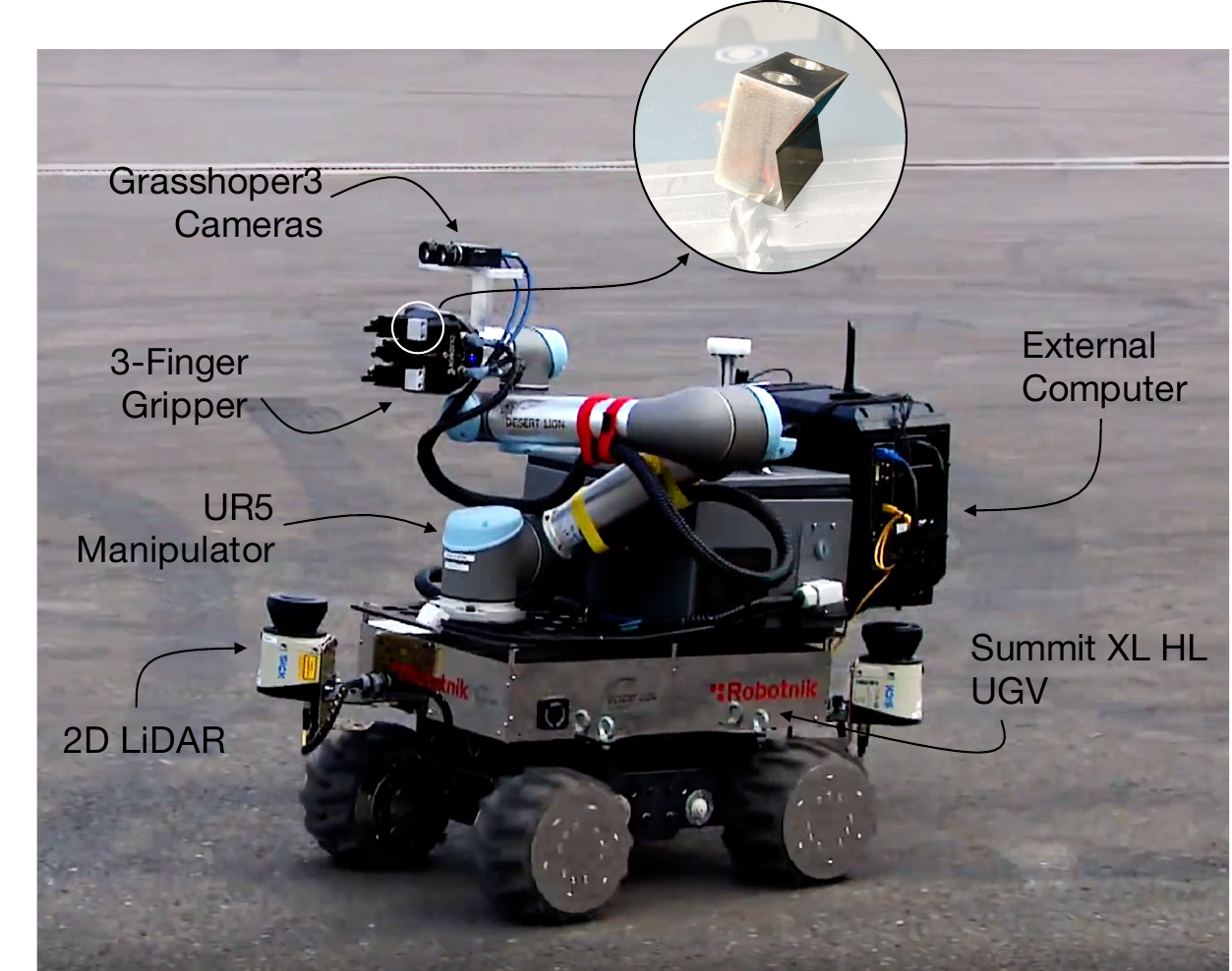} 
  \caption{The hardware configuration of RUR53 with a focus on the custom phalanges of the gripper, attached to improve its grasp capability.}\label{fig:robot_auto}
\end{figure}

RUR53 is a mobile manipulator composed of a Robotnik Summit XL HL
mobile base, a Universal Robot UR5
manipulator, and a Robotiq 3-Finger
adaptive gripper as depicted in Figure~\ref{fig:robot_auto}.
Thus, the robot can move within indoor and outdoor environments, reaching a speed of 10.8\,km/h, and it can manipulate lightweight objects of any dimension and geometry. To improve its grasp capability of the hand, we customized its phalanges by creating a groove able to canalize grasped tools (see the magnification of Figure~\ref{fig:robot_auto}).
Two SICK LMS151 outdoor 2D Light Detection And Ranging (LiDAR) sensors, one on its front and one on its back, let the robot navigate and explore its surroundings.
These sensors have an aperture angle of 270\degree, an angular resolution of 0.25\degree-0.5\degree, and an operating range of 0.5-50\,m, letting their combined usage sense almost the whole 60 $\times$ 60 m arena.
To recognize objects to be manipulated, a stereo vision system is mounted on a fixed support on the top of the gripper.
This system is composed of two Point Grey Grasshopper3 2.8MP Mono USB3 cameras located next to each other in Eye-on-Hand configuration to guarantee objects recognition even during the manipulation.
An embedded computer controls the mobile base while an external computer with a i7-6700 CPU, 16\,GB of RAM and a dedicated NVIDIA GTX 1070 perform heavy computational task.
The battery pack of the mobile base supplies energy to the entire RUR53.

\begin{figure}[t]
  \centering
  \includegraphics[width=0.9\linewidth]{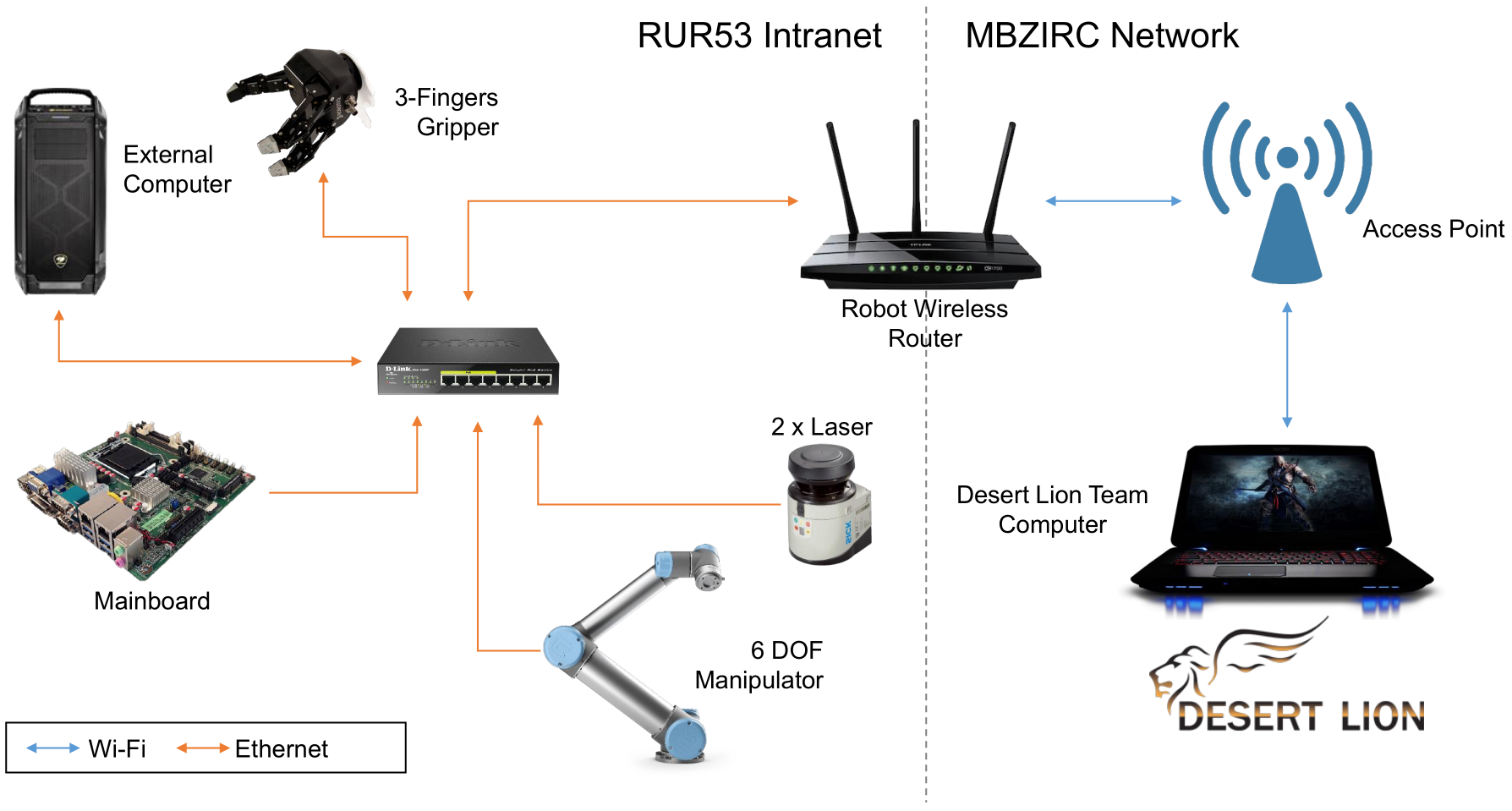} 
  \caption{RUR53 communication diagram.}\label{fig:communication}
\end{figure}

All robot components are connected through each other via Ethernet (orange lines of Figure~\ref{fig:communication}) and a WiFi router connects the robot with a external network, as the challenge arena network. In this way, a remote computer can control the robot and can receive feedback about its status.  
The router is characterized by two different WAN interfaces to enable a dual-channel connection: a wired one connected to the robot Intranet and a wireless one that can connect to an external network, like the MBZRIC one.
Given the limitations imposed by most of the router stock firmwares (e.g., the possibility of using Wireless interfaces as WAN interfaces), the router has been flashed with the last release of the open DD-WRT\footnote{\url{http://www.dd-wrt.com}} firmware.

\section{Software Architecture}\label{software}

The goal is letting RUR53 accomplish a set of complex tasks in an efficient way. Specifically to Challenge 2, these tasks require the outdoor navigation, the detection and recognition of some work tools, and their manipulation.
In order to face such complex tasks, the proposed software architecture has been optimized in order to be as modular as possible and let the user split the code and independently reuse its modules. 
Such modularity facilitates the redesign of individual elements, allowing a smart re-adaptation of the system to new workspaces and requirements, besides those of Challenge 2.

\begin{figure}[t]
  \centering
  \includegraphics[width =0.9\linewidth]{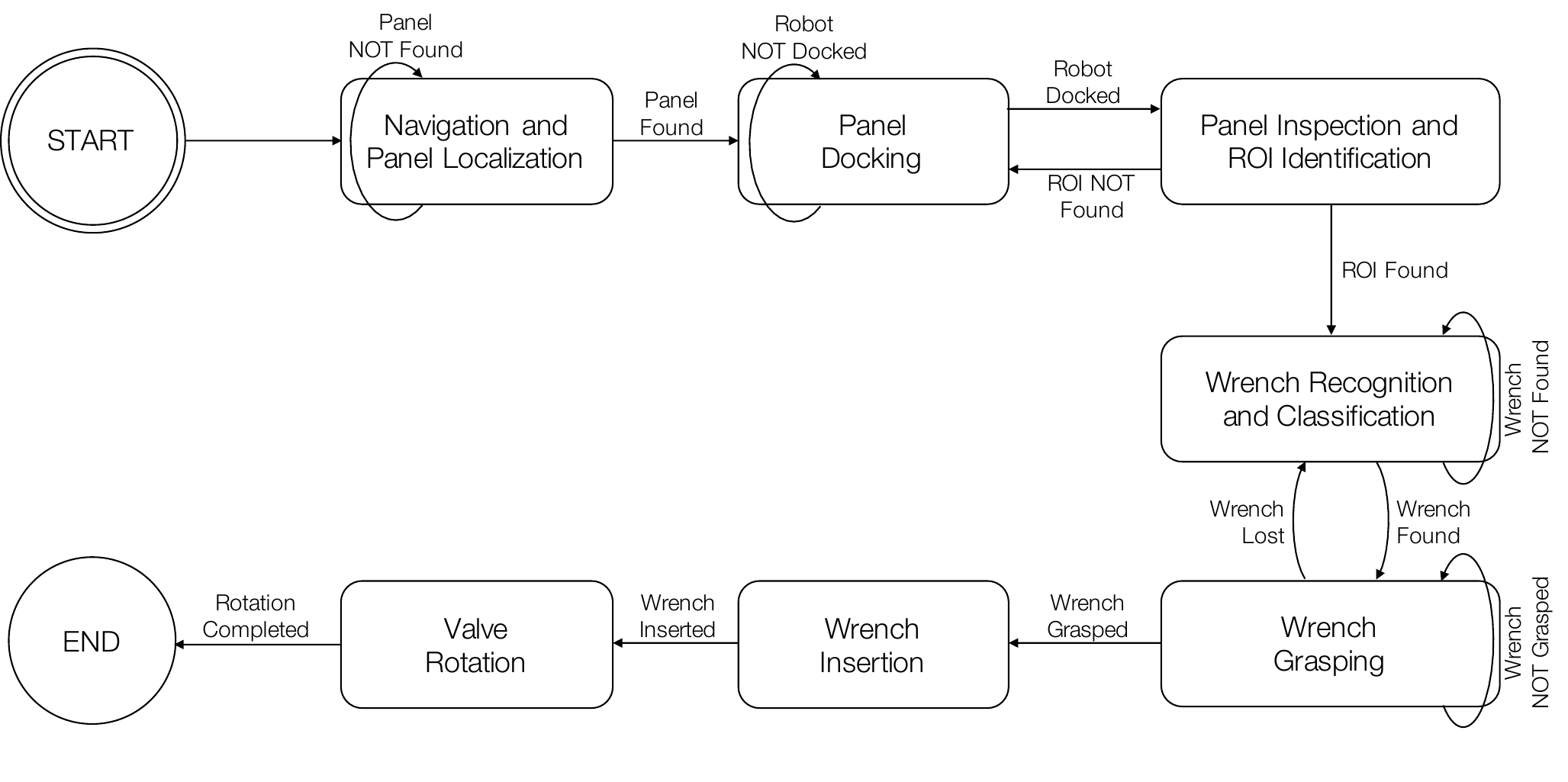} 
  \caption{RUR53 mission pipeline. Each block defines a task that the robot has to fulfill in order to complete the Challenge. A recovery plan has been designed in order to manage fault conditions, e.g, \textit{Panel NOT Found} or \textit{Wrench NOT Found}.}\label{fig:finite_state_machine}
\end{figure}

Figure~\ref{fig:finite_state_machine} depicts the proposed architecture, tailored to face Challenge 2 in an autonomous way. In detail, RUR53 navigates within the arena scanning for the panel. Once the panel is found, the robot approaches and inspects it looking for two Regions of Interest (ROI), one for the wrenches and one for the valve stem, respectively. If no ROI is found, RUR53 moves around the panel and inspects its other side. Otherwise, it examines the ROI in order to recognize and classify the right wrench. 
Once the wrench is found, RUR53 grasps and uses it to manipulate the valve stem. Recovery plans are designed to manage failures, such as \textit{Panel NOT Found}, \textit{Wrench NOT Found} or \textit{Wrench Lost} (see Figure~\ref{fig:finite_state_machine}). In the case of \textit{Wrench Lost}, for example, RUR53 repeats the wrench recognition and grasping steps in order to detect the other wrench suitable for the valve manipulation. Indeed, according to the challenge rules, two types of wrenches can operate the valve and the implemented recovery routine takes into consideration this information. Each of these modules is deeply described in the following sections.

\section{Navigation and Docking}\label{navigation}

\begin{figure}
  \centering
  \subfloat[]{
    \includegraphics[width = 0.4\textwidth]{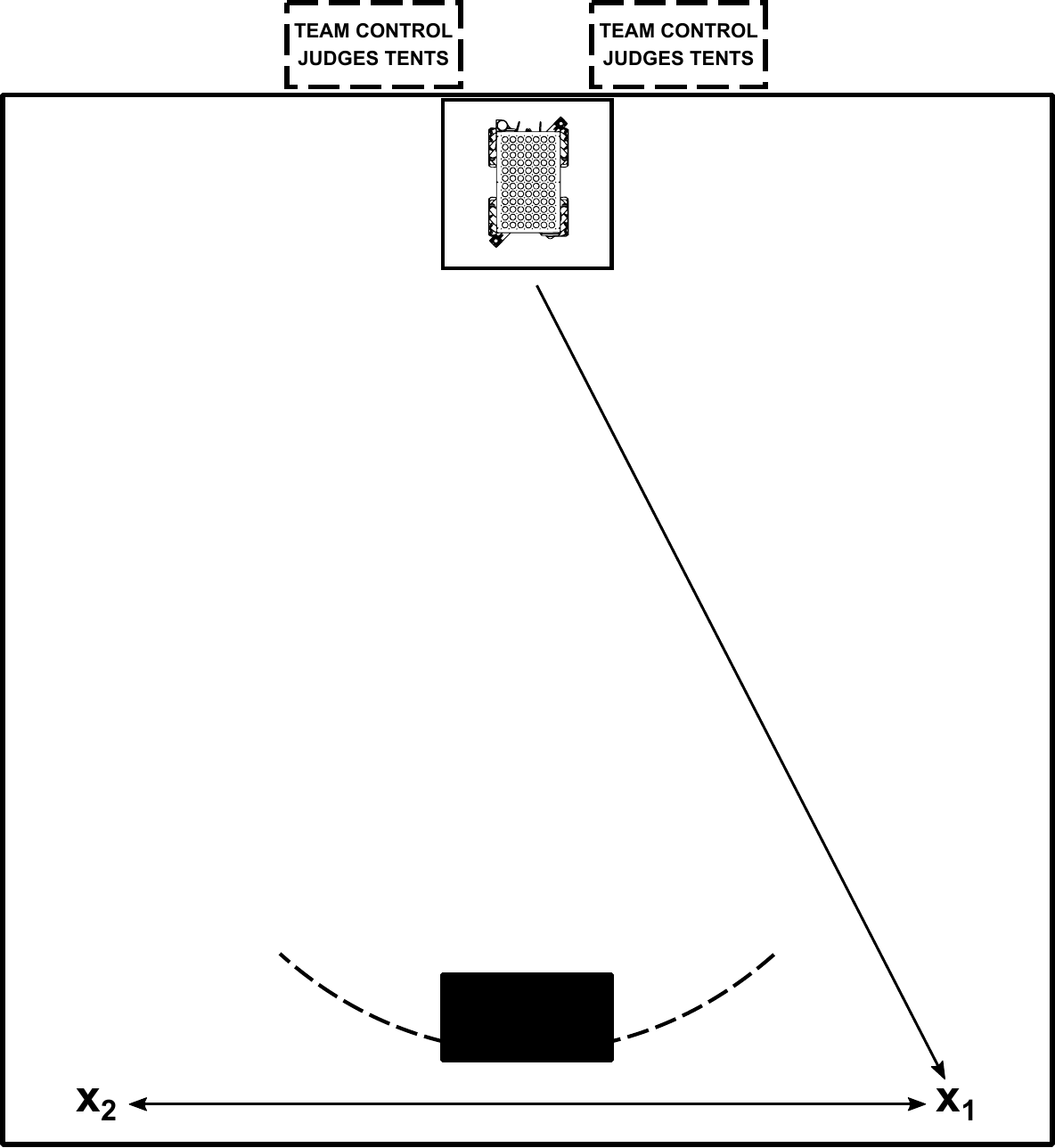}
  }
  \quad
  \subfloat[]{
    \includegraphics[width = 0.4\textwidth]{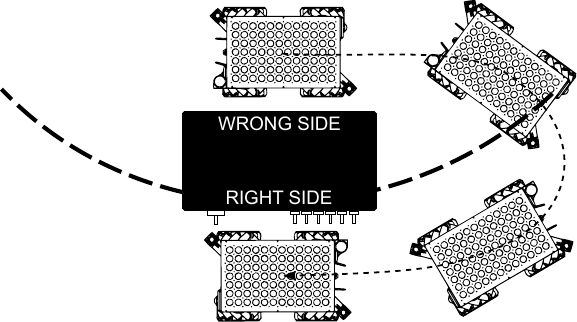} 
  }
  \caption{The navigation path in the challenge arena. The robot tries to reach the way-points $P = \{X_1, X_2$\}. They are visited periodically until the panel is found.}\label{fig:navigationpath}
\end{figure}

Challenge rules define the dimension of the arena and the curve along which the panel is randomly located (see Figure~\ref{fig:navigationpath}). Moreover, in the Grand Challenge, obstacles may exist (e.g., drones and landing vans).
Thus, the robot is programmed to navigate along a set of manually pre-defined waypoints $P = \{X_1, X_2$\}, periodically visited until the panel is found (see Figure \ref{fig:navigationpath}).
Between each couple of points, in order to prevent any collision, the implemented exploration routine exploits the ROS Navigation Stack, which provides collision avoidance, and laser-based SLAM (Simultaneous Localization and Mapping) through the GMapping algorithm~\cite{grisetti2005improving,grisetti2007improved}.

Dynamic Window Approach (DWA)~\cite{fox1997dynamic} and Timed Elastic Bands (TEB)~\cite{rosmann2012trajectory} are considered as local planners.
Both presents good performance in obstacle avoidance and planning accuracy, nevertheless TEB, without parallel planning, proves better performance than DWA.

During the arena exploration, the robot looks for the panel: the reference frames of the two lasers are semi-manually aligned by means of ICP~\cite{zhang1994iterative} while showing a known and fixed-sized object in the laser shared fields of view. Then, the two laser scans are merged taking the base link of the robot as reference.
From challenge rules, the size of the panel is known. Thus, the algorithm can search for its length through an euclidean clustering and a classification phase. In detail, given a set of clusters $C = \{C_i | 0 \leq i \leq N\}$ from the initial laser scan, it generates a subset $\widehat{C}$, such that $\widehat{C}\subseteq{C}$ and each $\widehat{C_j} \in \widehat{C}$ contains a set of points forming a line shape. 
This set is identified by fitting a line model on each cluster by means of the RANSAC algorithm~\cite{fischler1981random}. Clusters with too many outliers are not considered in $\widehat{C}$. $\widehat{C}$ is sorted so that $\widehat{C_0}$ collects the lengths closest to the panel size and $\widehat{C_N}$ stores the furthest ones.
To obtain this sorting, the algorithm exploits the size of the Oriented Bounding Box (OBB) of each cluster. Given the generic cluster $\widehat{C_k} = \{p_i \, | \, p_i \in \mathbb{R}^3\} \in {\widehat{C}}$, it computes the sizes $|\widehat{C_k}|_x$, $|\widehat{C_k}|_y$ and $|\widehat{C_k}|_z$ of the cluster's OBB by considering the minimum and maximum point of $\widehat{C_k}$ expressed in the reference system generated by the principal components $e_{k1}, e_{k2}, e_{k3}$ obtained from the covariance matrix $\Sigma_k$ of $\widehat{C_k}$. Let us assume N is the number of points $p_i$ in a generic cluster $\widehat{C_k} \in \widehat{C}$, then Equation~\ref{eq:OBBmeancov} defines the mean and the covariance matrix of $\widehat{C_k}$:
\begin{equation}
    \label{eq:OBBmeancov}
    \mu_k = \frac{1}{N}\sum_{i=0}^{N}p_i, \qquad \Sigma_k = \frac{1}{N - 1}\sum_{i = 0}^{N}(p_i - \mu_k)(p_i - \mu_k)^T\,.
\end{equation}
Equation~\ref{eq:OBBPCA} shows how to calculate the principal components of $\Sigma_k$:
\begin{equation}
    \label{eq:OBBPCA}
    \begin{aligned}
        \Sigma_k e_{ki} = \lambda_i
            e_{ki}, \qquad 1 \leq i \leq 3
    \end{aligned}\,.
\end{equation}
Equation~\ref{eq:OBBTransf} shows how to calculate the transformation $T$. This transformation brings each $p_i \in \widehat{C_k}$ from the original reference frame (i.e., the base link of the robot) to the new reference frame expressed by the principal components of $\widehat{C_k}$.
\begin{equation}
\label{eq:OBBTransf}
\begin{aligned}
T & = \begin{pmatrix}
            \horzbar & e_{k1}^T & \horzbar & \vertbar\\
            \horzbar & e_{k2}^T & \horzbar & \widehat{\mu_{k}}\\
            \horzbar & e_{k3}^T & \horzbar & \vertbar\\
            0        & 0        & 0        & 1
        \end{pmatrix}, 
        & \widehat{\mu_k} &= - \begin{pmatrix}
                            \horzbar & e_{k1}^T & \horzbar\\
                            \horzbar & e_{k2}^T & \horzbar\\
                            \horzbar & e_{k3}^T & \horzbar
                        \end{pmatrix} \mu_k \\
\end{aligned}\,.
\end{equation}
Finally, Equation~\ref{eq:sizesOBB} shows how to calculate the sizes of the final OBB of $\widehat{C_k}$, given the minimum and maximum points in the new reference system obtained by applying the transformation $T$ to each point in the cluster.
\begin{equation}
\label{eq:sizesOBB}
  \begin{matrix}
    |\widehat{C}_k|_x \\
    |\widehat{C}_k|_y \\  
    |\widehat{C}_k|_z
  \end{matrix}
  =
  \begin{pmatrix}
    |M_x - m_x| \\
    |M_y - m_y| \\
    |M_z - m_z|
    \end{pmatrix} 
\end{equation}
where
\begin{equation*}
  m  = \min\left(
    \widehat{p_i}
    \, \rule[-2ex]{0.5pt}{5ex}
    \, \begin{pmatrix}\widehat{p_i}\\
      0
    \end{pmatrix} = T \begin{pmatrix}p_i\\
      0
    \end{pmatrix}
    \right), \quad
    M = \max\left(
    \widehat{p_i}
    \, \rule[-2ex]{0.5pt}{5ex}
    \, \begin{pmatrix}\widehat{p_i}\\
      0
    \end{pmatrix} = T \begin{pmatrix}p_i\\
      0
    \end{pmatrix}
    \right)\,.
\end{equation*}
As the orientation of the panel is not known a priori, the algorithm sorts the dimensions of each $\widehat{C_k}$ and compares them to the ground truth panel dimensions by computing a similarity score: the cluster with the highest score represents the panel.
Thus, the robot can dock parallel to the panel. 
Once docked, the robot looks for the wrenches and, if it is not able to find them, the state manager moves the robot to change side (see Figure \ref{fig:navigationpath}). 

A perfect alignment between robot and panel is not guaranteed. Hence, the docking angle $\alpha$ between them is computed (see Figure \ref{fig:robot_orientation}). 
As the robot position with respect to the panel is approximately known, part of the laser points framing the panel can be easily retrieved by angle filtering. Then, Equation~\ref{line-plane} computes $\alpha'$, the docking angle between 0\degree and 90\degree: 
\begin{equation}\label{line-plane}
 \alpha' = \frac{Aa + Bb + Cc}{\sqrt{A^2+B^2+C^2}\sqrt{a^2+b^2+c^2}}\text{.}
\end{equation}
 where $(a, b, c)$ depicts the coefficients of the panel line and $(A, B, C, D)$ represents those of that robot side next to the panel. 
The dot product of the plane vector and the line vector provides the complementary to 90\degree. 
To provide an angle $\alpha$ in a $[0\degree-180\degree]$ range, the intersection between the panel line and the robot side plane has to be calculated. Then, the laser points should be projected to the robot side plane. This way, we can distinguish between $\alpha$ and $180\degree-\alpha$ by comparing the intersection point coordinates with the closest projected point coordinates.

\section{Perception}\label{perception}
Once docked, RUR53 select the right wrench and use it to operate the valve stem.
Both wrenches and valve are metallic and the challenge is outdoors, therefore non-negligible reflections and light variations follow.
Moreover, ISO~7738, cited by the Challenge's rules, defines only the dimension of the wrench head. Thus, the proposed approach exploits both 2D and 3D data of left and right camera images and reconstructs the point cloud through the Semi-Global Block Matching (SGBM)~\cite{Hirschmuller2008Stereo} algorithm\footnote{We use the implementation provided by OpenCV}.
Wrenches and valve stem are recognized by an object detector, while an accurate pose estimator retrieves their poses.

\subsection{Wrench and Valve Classifiers}\label{subsubsec:wrench_valve_classifier}
Two Cascade of Boosted Classifiers based on LBP features~\cite{Felzenszwalb2010Object} have been trained: one for the wrench and one for the valve detection.
During the training, the Hard Negative Mining~\cite{liao2007learning} technique has been exploited to add false positive and false negative samples, thus increasing the dataset.

Wrenches are classified based on the geometry of their heads. Indeed, as said before, ISO~7738 specifies the dimension of the head but the geometry and length of wrench handles may vary. Moreover, lab tests highlight that the head is always clearly visible and no obstruction exists affecting its recognition (see Figure \ref{fig:wrench_head_elaboration}). Pegs, instead, generate noise in both the 2D and 3D wrench reconstruction. 

\subsection{Wrench Detection and Pose Estimation}\label{wrench_detector}
Once located the wrenches, the manipulator aligns the vision system with the best point of view and starts finding the right wrench. 
Six different wrenches are randomly attached and only two of them can manipulate the valve stem. The proposed approach focuses on the detection of one assigned wrench, which exploits the images of the left camera together with the 3D point cloud of left and right cameras.

\begin{figure}[t]
\centering
  \includegraphics[width=0.95\textwidth]{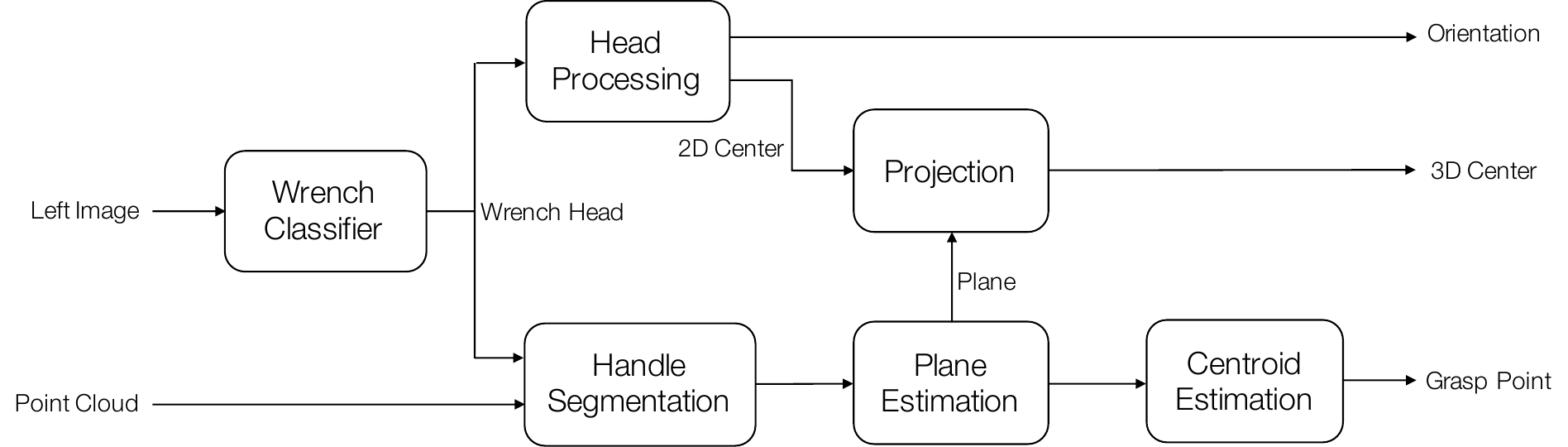} 
  \caption{Wrench processing pipeline. The pipeline takes as input the left image and the point cloud (i.e., the image detected by the left camera). It estimates the center and orientation of the wrench head and the wrench grasp point.}
  \label{fig:wrench_detection_overview}
\end{figure}

\begin{figure}[t]
\centering
\subfloat[\label{fig:wrench_classifier}]{%
  \includegraphics[width=0.38\linewidth]{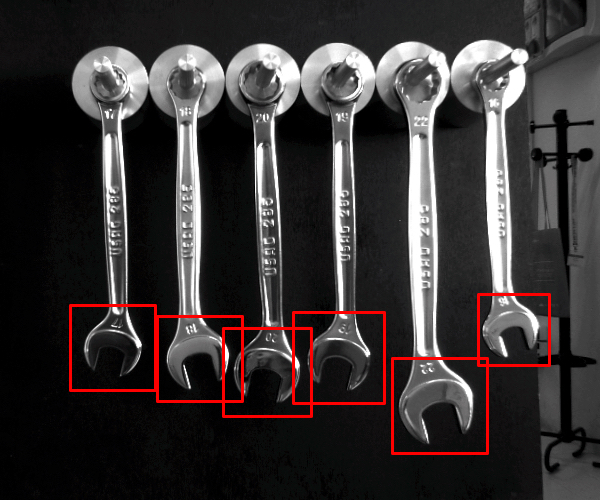}%
}
\quad
\subfloat[\label{fig:wrench_head_elaboration}]{%
    \includegraphics[width=0.15\linewidth]{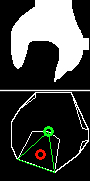}
}
\quad
\subfloat[\label{fig:wrench_result}]{%
  \includegraphics[width=0.38\linewidth]{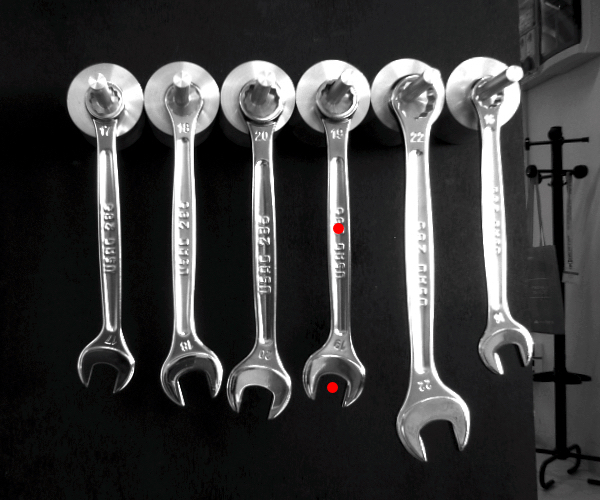}%
}
\caption{Wrench detection, recognition, and grasp point estimation processes. (a) The result of the detection routine: wrench heads are surrounded by \textit{red} squares.
(b) The wrench recognition processing: the image of the wrench head has been binarized (\textit{upper} figure); the convex-contour of the head is computed (\textit{white} line of the \textit{lower} figure) and its convexity defect is retrieved (\textit{green} triangle) together with its maximum point (\textit{green} circle). The grip center of the head is obtained as the centroid of the green triangle (\textit{red} circle). (c) The point where the wrench has to be grasped together with its grip center (\textit{red} points).
}\label{fig:wrench_elaboration}
\end{figure}

\begin{figure}[t]
\centering
\subfloat[Front view.]{%
  \includegraphics[width=0.43\linewidth]{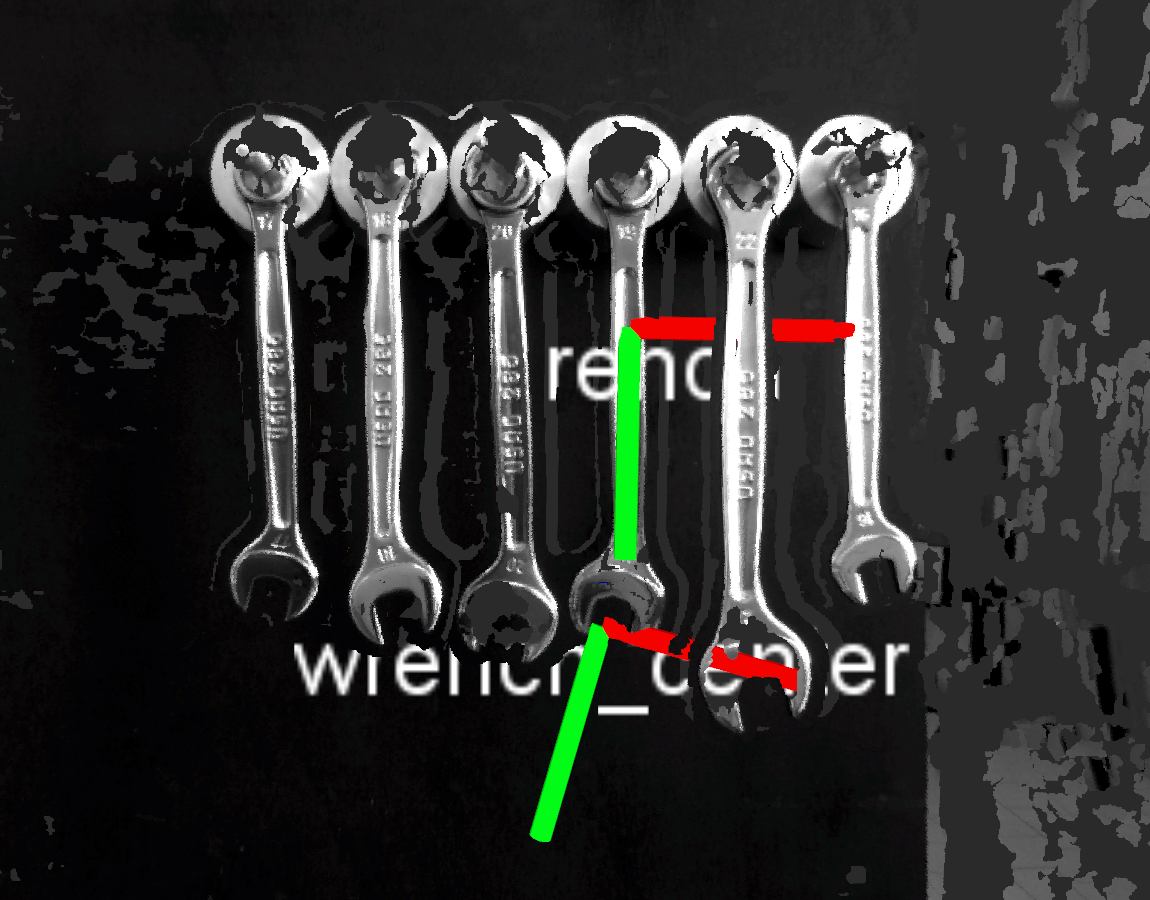} 
}
\quad
\subfloat[Side view.]{%
  \includegraphics[width=0.43\linewidth]{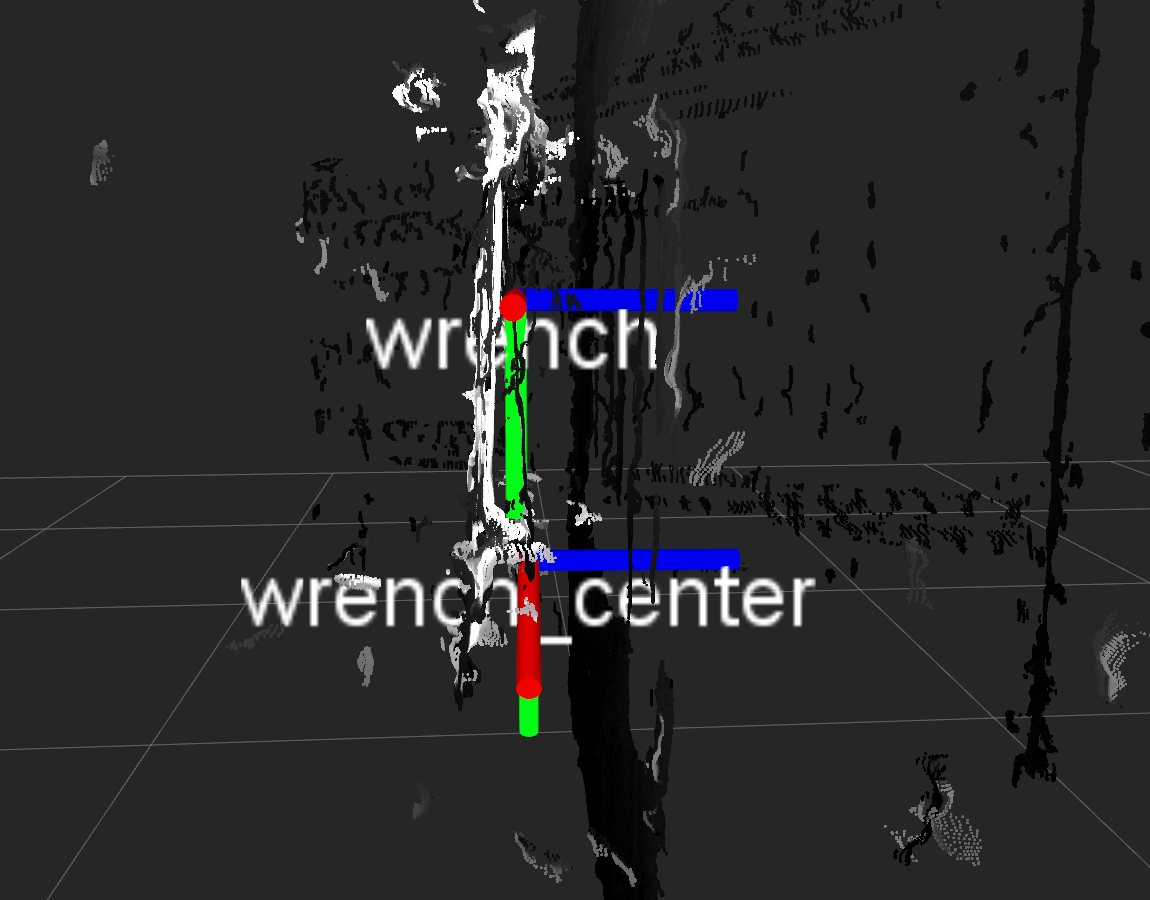} 
}
 \caption{Wrenches orientation axes: \textit{(green, red, blue)} colors represent \textit{(x, y, z)} axes, respectively.}
 \label{fig:wrench_3d}
\end{figure}

First, the wrench detector finds the wrenches heads and provides a ROI for each of them (see Figure~\ref{fig:wrench_classifier}).
For each ROI, the estimation of the grip center of the head, the wrench orientation, and the handle centroid (see Figures~\ref{fig:wrench_result} and \ref{fig:wrench_3d}) follow.
In order to be robust to wind, reflections, noise, and detection outliers, each of these values is computed as the median of single-view values over 10 frames through two simultaneous steps (see Figure~\ref{fig:wrench_detection_overview}). 

The former estimates the grip center of the wrench head. The latter finds the grasping point on the wrench handle: first, the handle is included by extending the ROI along its vertical axis; then, the region of the grasp point is segmented on the resulting point cloud. To do that, all the points closer than 1~m from the camera are extracted. Moreover, all the points belonging to the panel or other outliers are removed by computing the mean distance $\bar{z} = \frac{1}{n} \sum_{i=0}^{n} z_i$  of the points from the camera, where $n$ is the number of points in the cluster and $z$ is the coordinate perpendicular to the plane of the camera. 
All the points farther than 1.5\,cm from the mean value (this value being experimentally chosen) are removed from the cluster.
Finally, the plane $\pi$ of the wrench handle is estimated by fitting on it a plane modeled via RANSAC~\cite{fischler1981random}. All points fitting the model must have a maximum Euclidean distance of 1\,cm from the plane; outliers are then discarded.
Once the segmentation of the point cloud of the handle is achieved, the grasp point is computed as the centroid of these points. The $z$ coordinate of the grasp point is considered perpendicular to the panel. As described in Section~\ref{navigation}, the panel orientation is estimated after the docking.
 
This pose gives information about the orientation of the wrench itself. Indeed, the head can be turned to right or left and, depending on it, the gripper has to change how to manipulate the wrench in order to fit it into the valve stem (in Figure~\ref{fig:wrench_head_elaboration}, the head is oriented to the left). The other step finds the grasping point on the wrench handle. In this case, the input ROI is extended along its vertical axis in order to include the handle. Thus, if the detected head bounding box is $\mathit{bbox}_{\mathit{head}} = (x, y, \mathit{width}, \mathit{height})$, the bounding box of the handle will be $\mathit{bbox}_{\mathit{handle}} = (x, y-2*\mathit{height}, \mathit{width}, 2*\mathit{height})$. 
From this handle bounding box, the grasp point is computed by segmenting the corresponding region on the point cloud. As a first step, all the points with a maximum distance of 1\,m from the camera are extracted. 
Then, in order to remove points belonging to the panel or other outliers, the mean distance $\bar{z} = \frac{1}{n} \sum_{i=0}^{n} z_i$ of the points from the camera is calculated,

The wrench head orientation needs to be accurately evaluated. Indeed, the head can be turned to right or left and this orientation affects the procedure that joints wrench and valve (in Figure~\ref{fig:wrench_head_elaboration}, the head is oriented to the left). 
First, the ROI of the head is binarized by using the Otsu's thresholding algorithm~\cite{otsu1979threshold} (Figure~\ref{fig:wrench_head_elaboration}, \textit{upper} figure). Then, the contours of the binarized image are retrieved by~\cite{suzuki1985topological} and the convex hull of the contours (Convex-contour) is found by using the Sklansky’s algorithm~\cite{sklansky1982finding} (Figure~\ref{fig:wrench_head_elaboration}, \textit{lower} figure, \textit{white} contour). The largest convexity defect of the retrieved contour is computed (Figure~\ref{fig:wrench_head_elaboration}, \textit{green} triangle) and its centroid (Figure~\ref{fig:wrench_head_elaboration}, \textit{red} circle) reflects the head's grip center. It is characterized by the 2D coordinates $(u_c, v_c)$, while the \textit{x}- and \textit{y}- orientations (Figure \ref{fig:wrench_3d}, \textit{green} and \textit{red} lines) are derived from the angle of the line passing through the maximum point of the convexity defect (Figure~\ref{fig:wrench_head_elaboration}, \textit{green} circle) and the grip center itself.

Equation~\ref{eq:wrench_inclination} computes the head orientation:
\begin{equation}
\label{eq:wrench_inclination}
\beta = \arctan\left(\frac{v_d-v_c}{u_d-u_c}\right)
\text{,}
\end{equation}
where $(u_d, v_d)$ are the coordinates of the deep point (Figure \ref{fig:wrench_head_elaboration}, \textit{green} circle) and $(u_c, v_c)$ are those of the center point (Figure \ref{fig:wrench_head_elaboration}, \textit{red} circle). The \textit{z}-orientation, as for the handle of the wrench, is considered perpendicular to the panel (Figure \ref{fig:wrench_3d}, \textit{blue} line).

\begin{figure}[t]
\centering
  \includegraphics[width=0.6\linewidth]{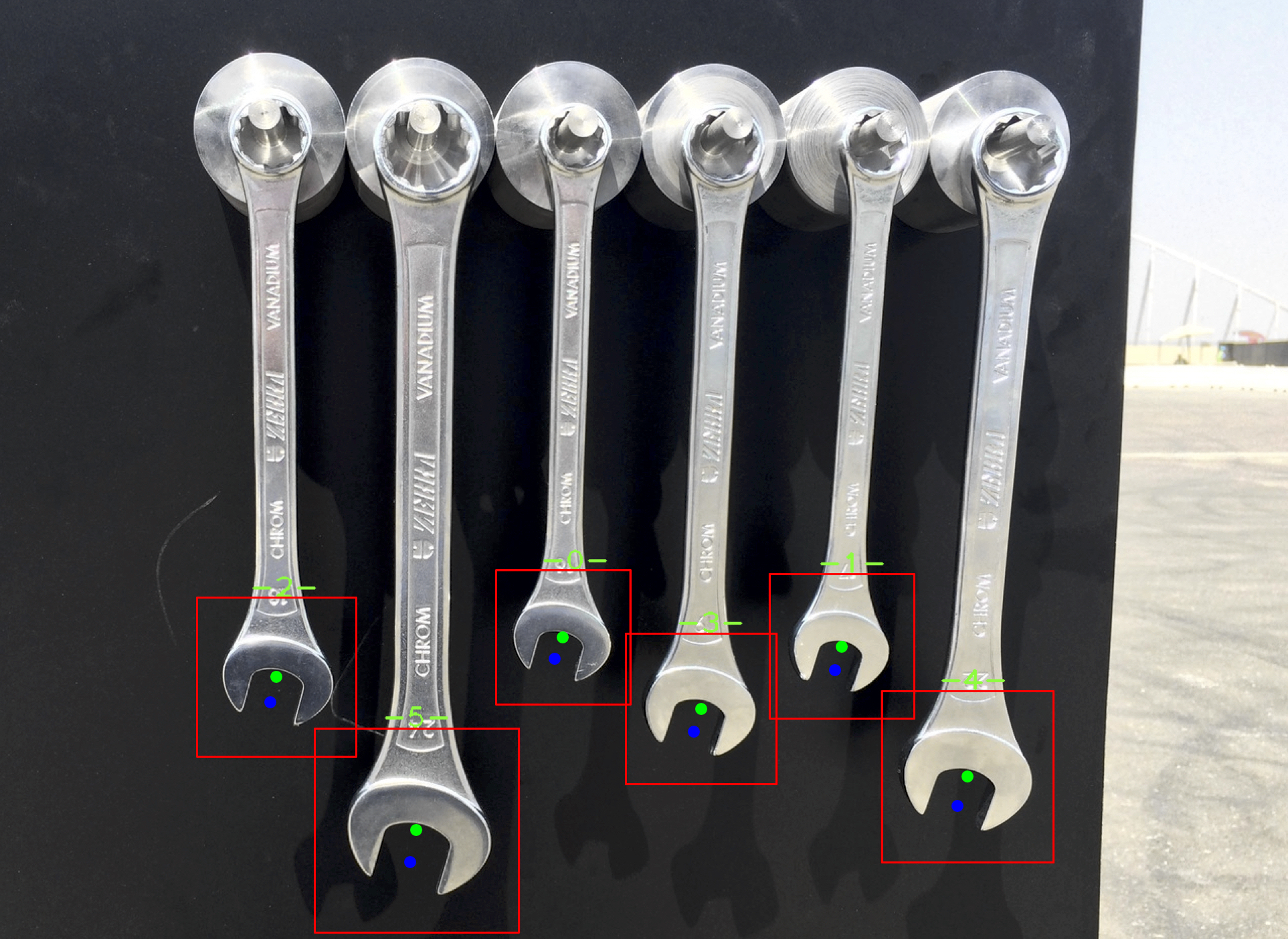}%
\caption{Wrench classification on the Challenge panel.}\label{fig:mbzirc_detection}
\end{figure}

Finally, to find the 3D coordinates of the grip center, the 2D coordinates of the center $(u_c, v_c)$ are projected onto the plane $\pi$: the 3D ROI (in the world coordinate system) of the grip center is extracted from its 2D coordinate. Then, Equation~\ref{eq:plane_1} and Equation~\ref{eq:plane_2} compute the intersection of the ray with $\pi$:
\begin{equation}
\label{eq:plane_1}
t = -D / (A * x_{\mathit{ray}} + B * y_{\mathit{ray}} + C)\,,
\end{equation}
\begin{equation}
\label{eq:plane_2}
\begin{cases}
x_c = t * x_{\mathit{ray}}\\
y_c = t * y_{\mathit{ray}}\\
z_c = t
\end{cases}
\text{,}
\end{equation}
where $(A, B, C, D)$ are the coefficients of the plane $\pi$ and $(x_{\mathit{ray}}, y_{\mathit{ray}})$ are the coefficients of the 3D ray.
Figure~\ref{fig:mbzirc_detection} shows the result obtained on the panel of the Challenge.

\subsection{Valve Detection and Pose Estimation}\label{valve_detector}

\begin{figure}
\centering
    \includegraphics[width=0.95\linewidth]{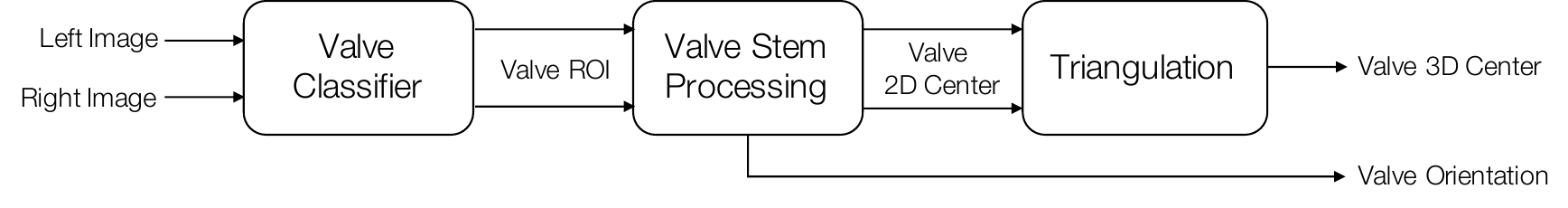} 
    \caption{Valve processing pipeline. The pipeline receives the left and right images as input and estimates the center and orientation of the valve stem.}
    \label{fig:valve_detection}
\end{figure}

\begin{figure}[t]
\centering
\subfloat[\label{fig:valve_step1}]{%
    \includegraphics[width=0.27\linewidth]{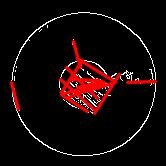} 
}
\quad
\subfloat[\label{fig:valve_step2}]{%
    \includegraphics[width=0.27\linewidth]{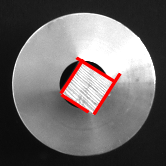} 
}
\quad
\subfloat[\label{fig:valve_step3}]{%
    \includegraphics[width=0.27\linewidth]{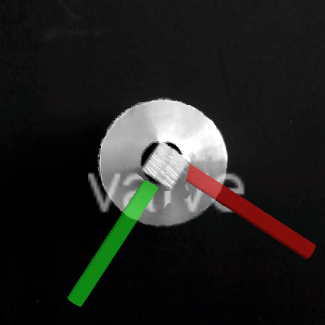} 
}
\caption{Valve detection and pose estimation. (a) Contours extracted from the initial image. (b) The best combination of 4 line segments that respects the constraints. (c) The estimated position and orientation of the valve stem.}\label{fig:valve_steps}
\end{figure}

\begin{figure}[t]
\centering
\includegraphics[width=0.45\linewidth]{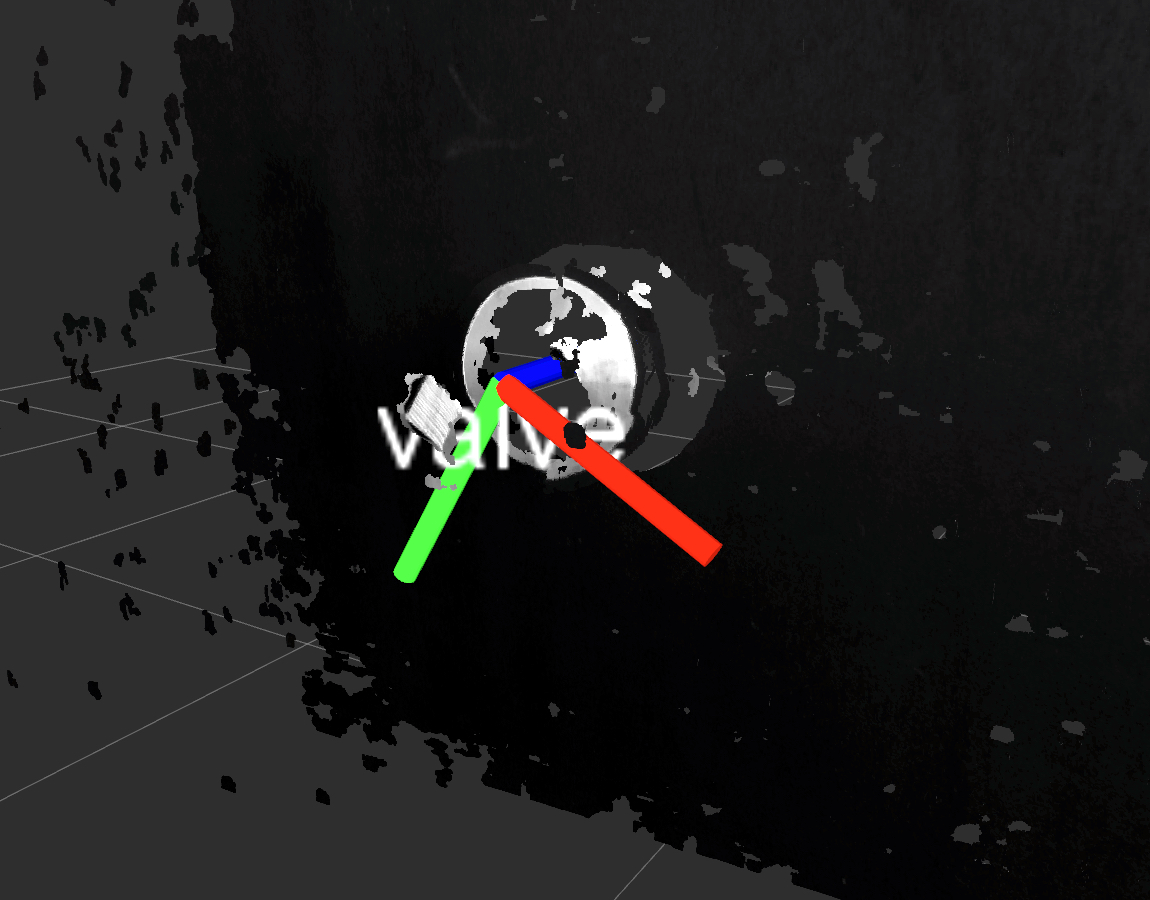} 
\caption{Estimated valve stem position and orientation on the panel used during the Challenge. \textit{(x, y, z)} axes are colored of \textit{(red, green, blue)}, respectively.}\label{fig:valve_3d}
\end{figure}
Once the right wrench is grasped, the valve stem should be located so as to be able to manipulate it. Being the valve metallic and reflective, its 3D reconstruction is noisy and useless. Thus, the Triangulation~\cite{hartley1997triangulation} algorithm is applied on the points of interest extracted from the stereo images (\textit{i.e.}, left and right images), principally on those depicting the center of the valve.

The proposed approach is shown in Figure~\ref{fig:valve_detection}). In detail, as the distance between wrenches and valve is known and fixed, the arm automatically moves on the valve stem in order to let the vision system frame it. Positioning the cameras in from of the valve considerably reduces noise and shadows due to reflections. Then, the valve detector extracts the ROI of the valve and the Canny Edge Detector~\cite{canny1986computational} uses it to extract the valve stem edges. All line segments that may belong to the front face of the stem are extracted through the Probabilistic Hough Transform~\cite{matas2000robust} (see Figure~\ref{fig:valve_step1}).
Given the set of lines and knowing that the valve is a square of known dimensions (from the Challenge rules), the procedure searches for those four perpendicular line segments which define the front face of the valve itself. First, those line segments that are too short or too long are removed. Then, all possible combinations of four line segments are generated and for each combination, the following checks are performed: i) there exists an end point of a line segment close to the end point of another line segment (only in this case their combination forms a vertex of the square); ii) there exists a line segment parallel to another line segment; iii) there exists a line segment perpendicular to another line segment.
If none of these combinations exists, the check is repeated on all possible groups of three segments (see Figure~\ref{fig:valve_step2}).

Once the candidate line segments are found, a square is formed and its center is the center of the valve stem. Equation~\ref{eq:inclination} computes the orientation angle normalized in  [$0\degree$ - 90\degree]. The \textit{z}-orientation of the valve, instead, is considered perpendicular to the panel, whose orientation is estimated after the docking routine of Section~\ref{navigation}.
\begin{equation}
\label{eq:inclination}
\gamma = \arctan\left(\frac{y}{x}\right)
\end{equation}
Finally, the  Hartley's algorithm~\cite{hartley1997triangulation} is applied to triangulate left and right data and obtain the final 3D position of the valve center (see Figure~\ref{fig:valve_step3}).  Figure~\ref{fig:valve_3d} shows the result obtained during the Challenge.

\section{Manipulation}\label{manipulation}

\begin{figure}
 \centering
\subfloat[\label{fig:sfig1}]{
  \includegraphics[width=0.45\linewidth]{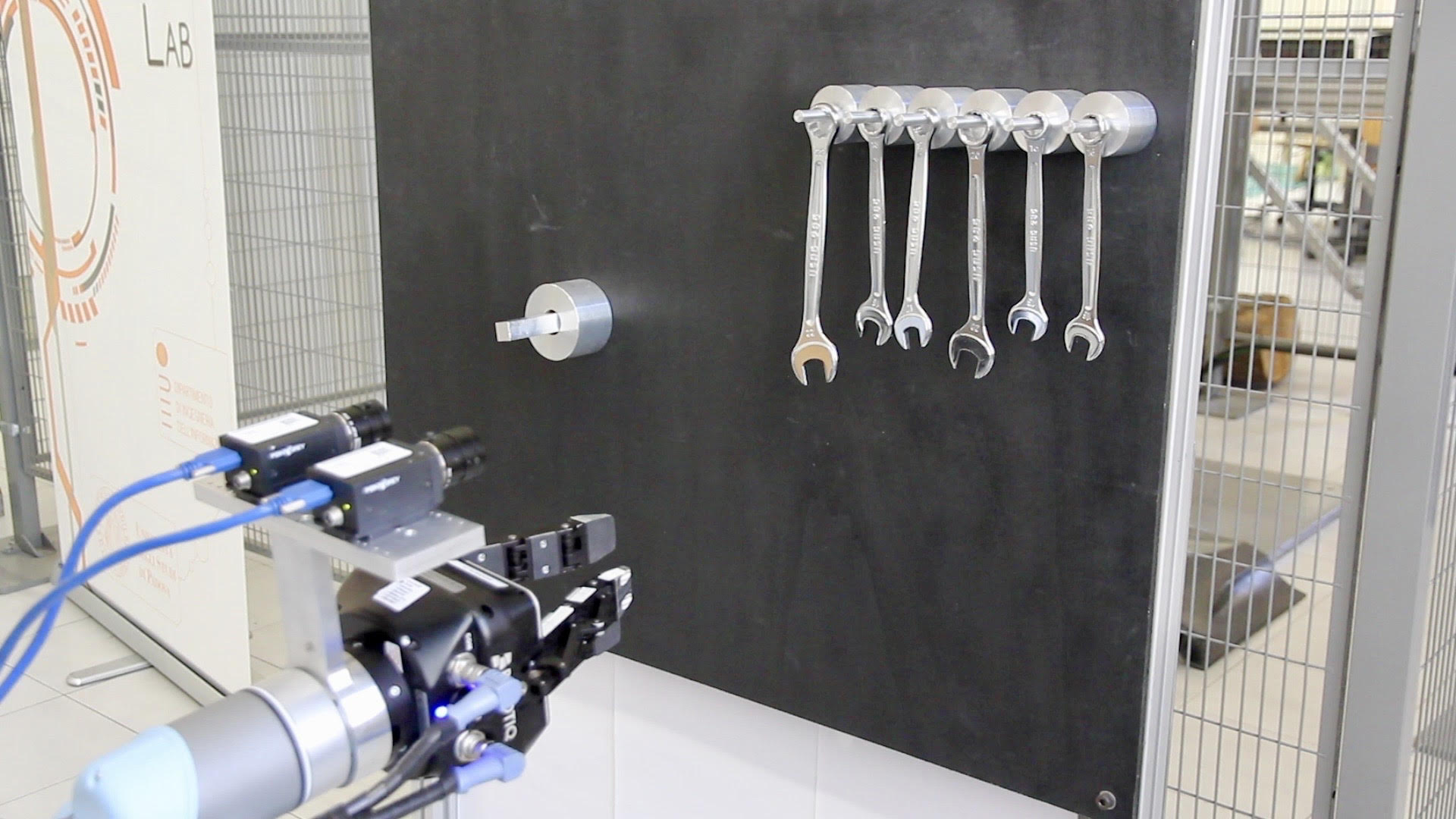}
}
\subfloat[\label{fig:sfig2}]{
  \includegraphics[width=0.45\linewidth]{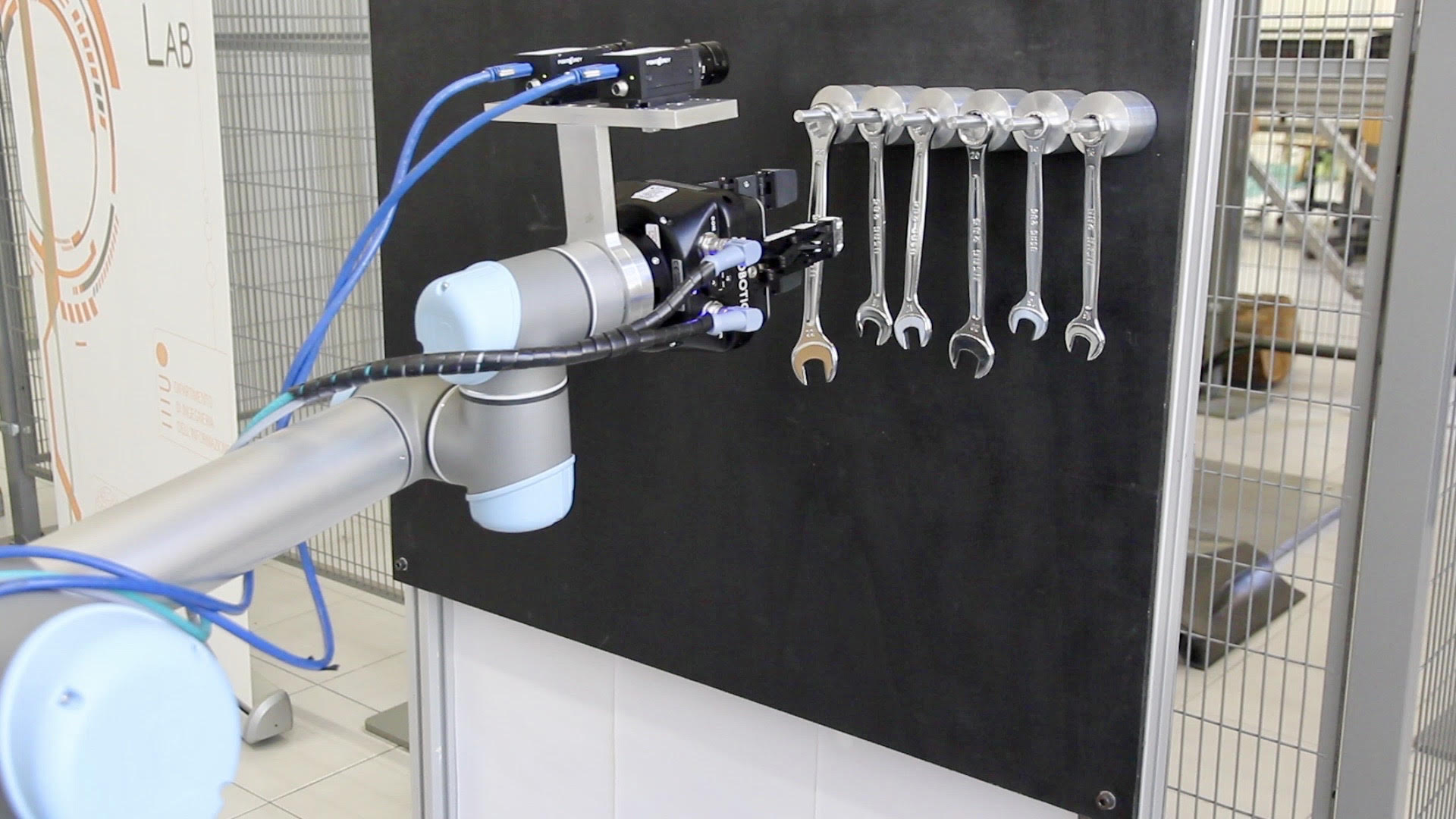}
}\\
\subfloat[\label{fig:sfig3}]{
  \includegraphics[width=0.45\linewidth]{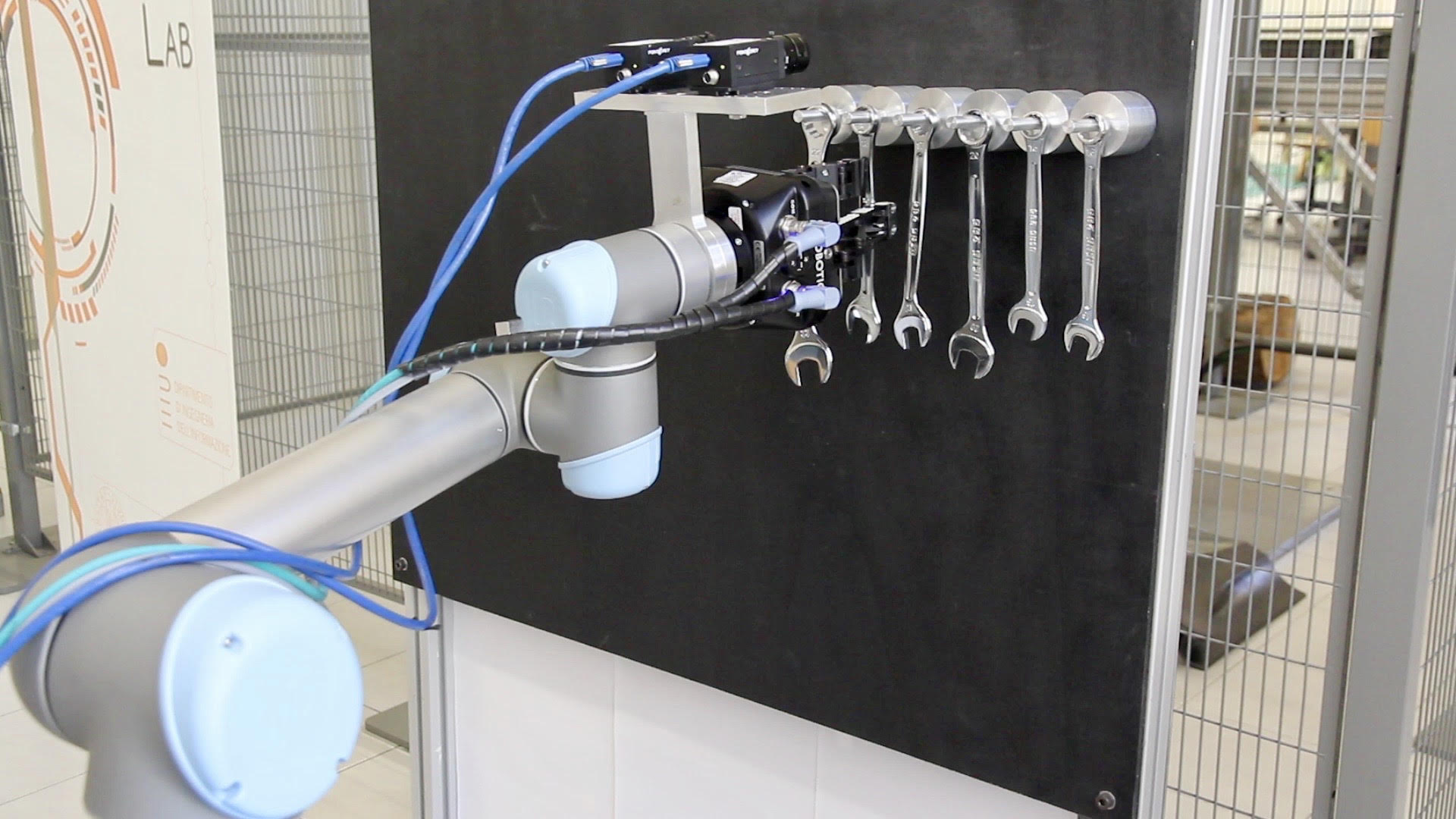}
}
\subfloat[\label{fig:sfig4}]{
  \includegraphics[width=0.45\linewidth]{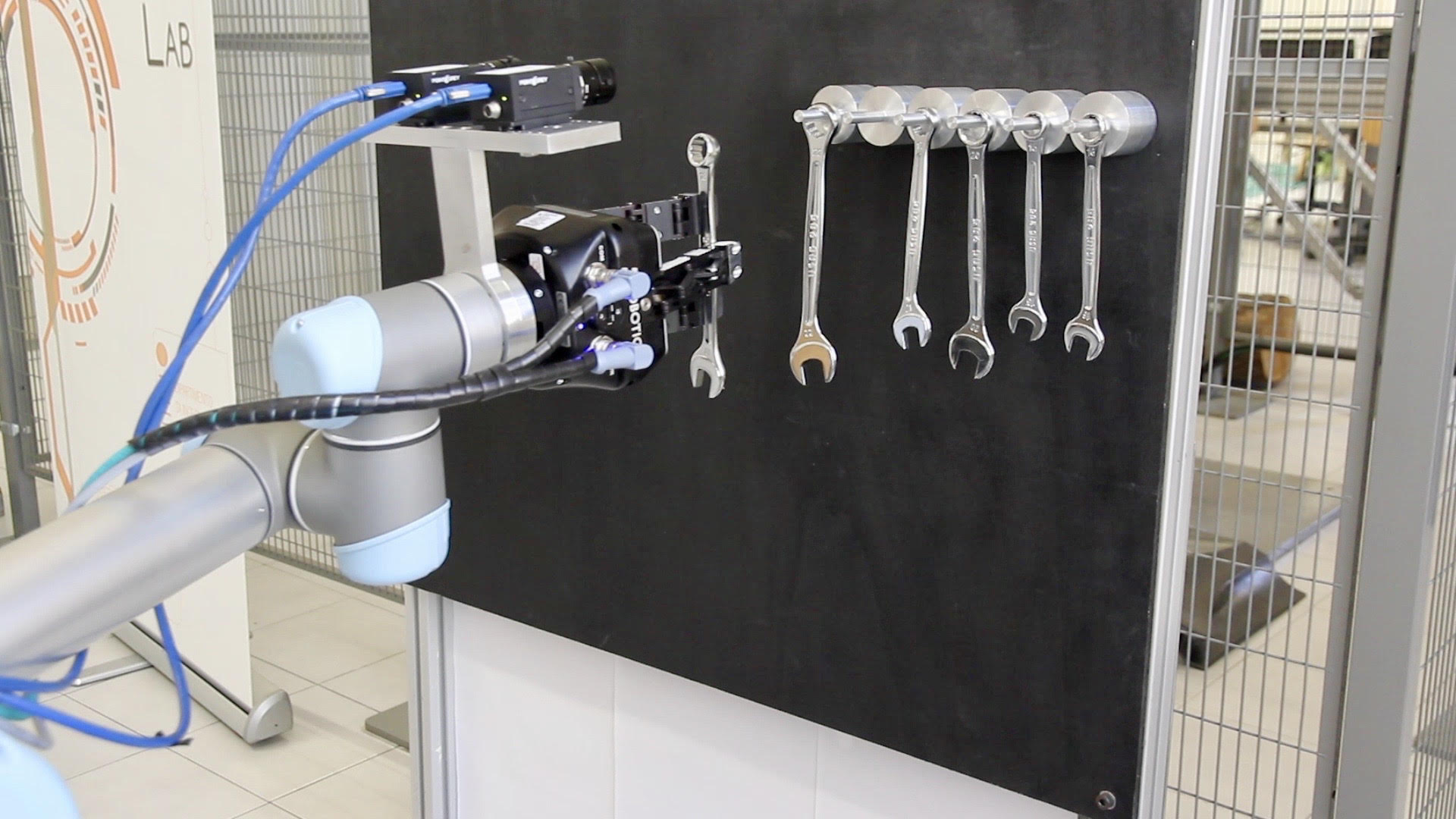}
}
\caption{The autonomous wrench grasping routine. (a) The vision system detects the right wrench. (b) The arm approaches the panel. (c) The gripper grasps the correct wrench. (d) The wrench is extracted.}
\label{fig:grasp}
\end{figure} 

Once the wrench area is detected (Section \ref{wrench_detector}), the manipulator moves to align the end-effector cameras with the detected ROI. 
Once the right tool is recognized and its grasp point is computed (Section \ref{wrench_detector}), two oriented reference systems are defined: one on the grasp point and one on the grip center of the wrench head (see Figure \ref{fig:wrench_3d}). The end-effector is aligned with the grasp point (fingers opened), it approaches the wrench, closes the fingers, and comes back to its initial pose (see Figure \ref{fig:grasp}).

Finally, since the valve position is known, the end-effector moves in front of it and the detection module computes its pose (Section ~\ref{valve_detector}). An oriented reference system is placed on the center of the valve (see Figure \ref{fig:valve_3d}) and the reference systems of valve and wrench grip centers are aligned. 
The robot inserts the wrench on the valve rotates it $360\degree$ by sampling a set of waypoints on a radius $r$ around the valve, with $r$ the distance between the center of the valve (equal to the wrench grip center) and the wrench grasp point.

To perform the manipulation tasks, the UR5 is controlled in its tool-space using the URScript, that is sent to the robot through the UR Modern Driver ROS package. The gripper, instead, is controlled through the Robotiq ROS package.

\section{Teleoperation}\label{teleoperation}

\begin{figure}
 \centering
  \includegraphics[width = 0.6\textwidth]{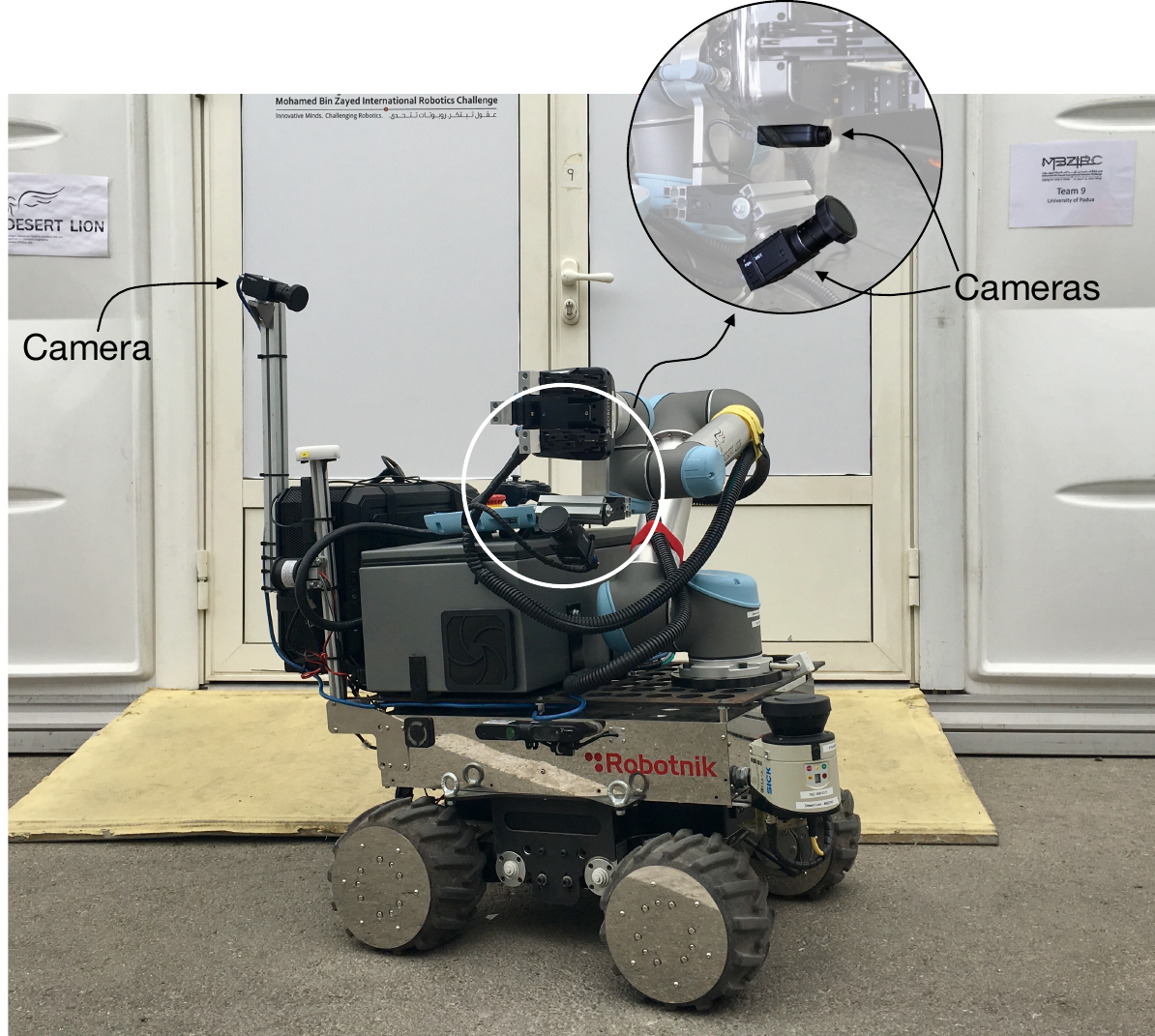} 
 \caption{RUR53 hardware configuration in the teleoperated mode and a zoom-in on the gripper's cameras.}
 \label{fig:robot_teleop}
\end{figure}

In the teleoperation mode, the robot configuration changes from Figure~\ref{fig:robot_auto} to Figure~\ref{fig:robot_teleop} to let the pilot safety drive the robot even when far from the cockpit. This new configuration provides one camera mounted on the mobile base and two cameras mounted under the gripper.
The first camera gives a panoramic view of the robot surroundings and lets check the pose of the mobile base with respect to the panel, the pose of the wrench within the gripper, and its alignment with respect to the valve stem. 
Cameras under the gripper let find the right wrench, verify the correctness of the grasp, and check the joint of wrench and valve stem.

\begin{figure}[t]
 \centering
\subfloat[\label{fig:wrench_to_gripper}]{
  \includegraphics[height=6cm]{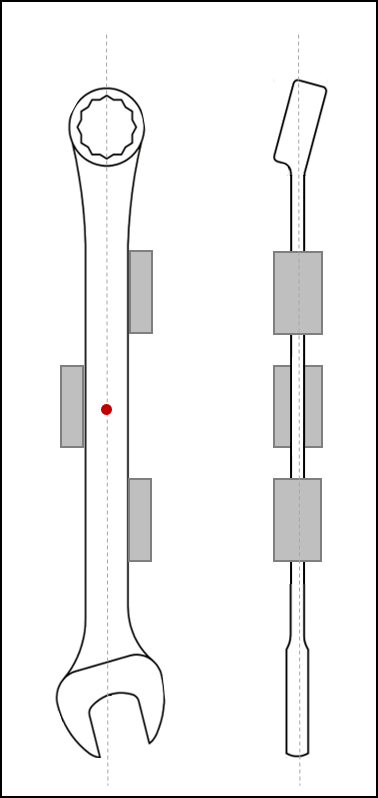}
}
\subfloat[\label{fig:wrench_to_valve}]{
  \includegraphics[height=6cm]{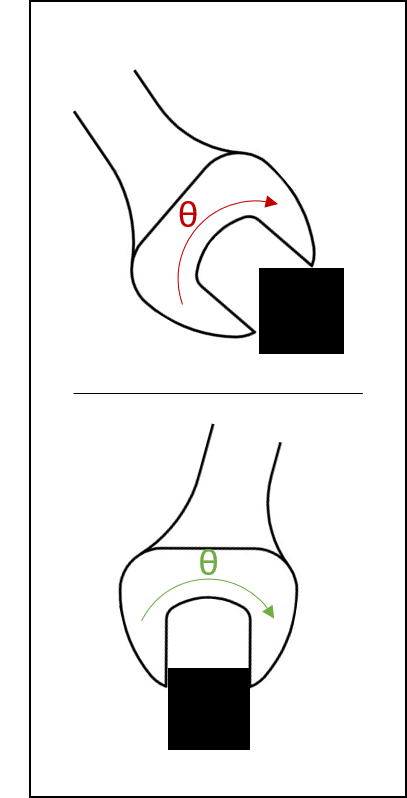}
}
\caption{Wrench orientations necessary for a correct valve rotation. (a) The orientation of the wrench with respect to the gripper fingers (\textit{red}: the computed grasp point, \textit{gray}: the fingers). (b) The orientation of the wrench with respect to the valve (\textit{red}: an initial wrong alignment implies an inaccurate valve rotation; \textit{green}: an initial exact alignment facilitates a correct 360$\degree$ valve rotation). }
\label{fig:right_movements}
\end{figure}

The proposed routine lets teleoperate the robot by imprinting velocity commands to each of its components (mobile base, arm, and gripper) through the keyboard. Thus, the pilot can send velocity commands to the Summit XL HL (e.g., go forward, go backward, turn left and turn right); via different keyboard keys, she/he can move and rotate the UR5 end effector as well as opening and closing the gripper. 
Once the robot attaches the wrench to the valve, its $360\degree$ rotation is performed by imprinting successive small rotations to the end-effector, calculated from the position of the wrench on the valve and the point in which the wrench is grasped.
To avoid the loss of the wrench (see Figure~\ref{fig:wrench_to_gripper}) or its detachment during the rotation (Figure~\ref{fig:wrench_to_valve}, \textit{upper} figure), successive $5\degree$ rotations are imprinted when the wrench is not firmly grasped or when it does not firmly fit the valve. 
$15 \degree$ rotations, instead, are imprinted when the grasp is safe enough (Figure~\ref{fig:wrench_to_valve}, \textit{lower} figure).

\section{Experimental Results}\label{results_discussion}
Proposed experiments aim to highlight the performance of the robot in fully autonomous mode. Experiments have been performed both in our laboratory and during the challenge. 
As explained in Section \ref{lessons_learned}, during the challenge, the detection of the panel was not possible. Thus, experiments performed during the challenge concern only the perception routine.

\subsection{Panel detection and Docking}
The implemented panel detection algorithm aims to identify and estimate the pose of the panel while exploring the arena. The algorithm lets the robot perform the detection even when obstacles occlude the path but no other object should exists with a geometry similar to that of the panel. 

\begin{figure}[t]
  \centering
  \subfloat[\label{fig:8m}]{
    \includegraphics[width=0.45\textwidth]{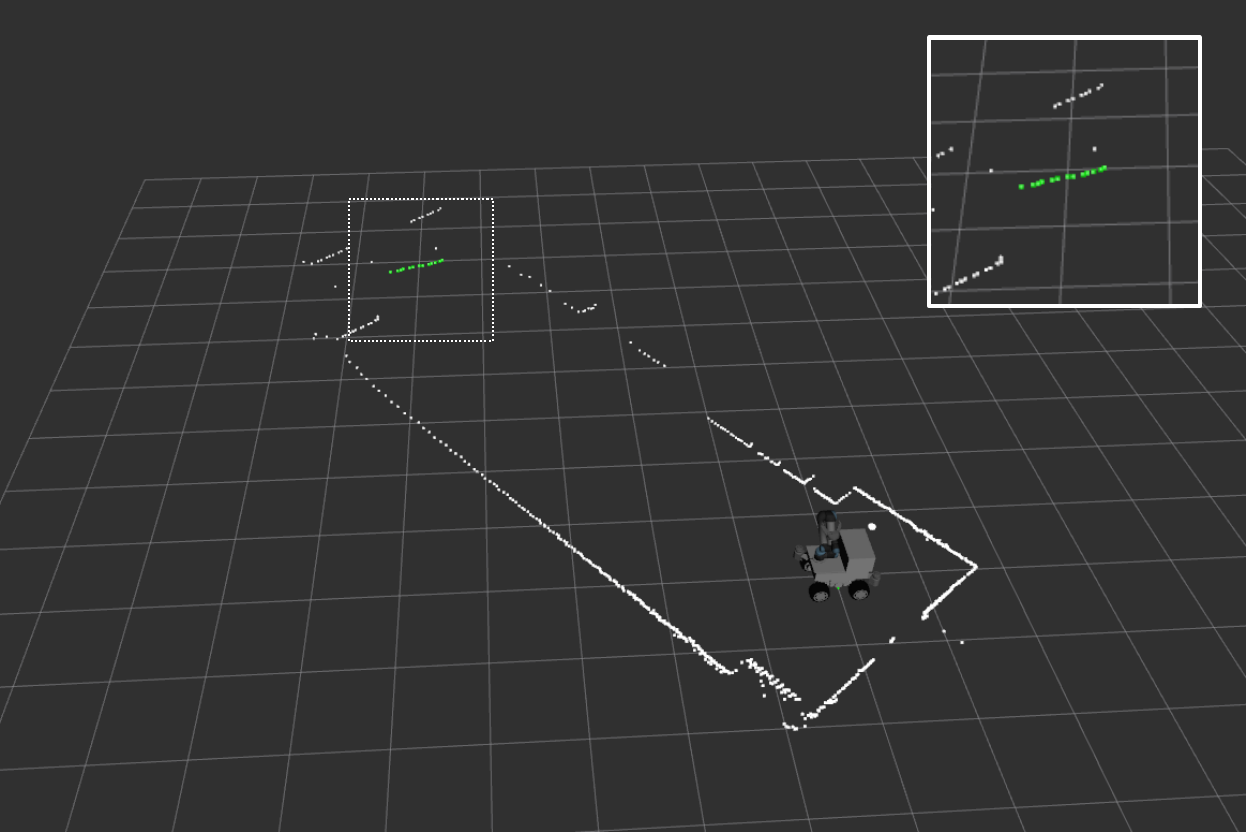}
  }
  \subfloat[\label{fig:5m}]{
    \includegraphics[width=0.45\textwidth]{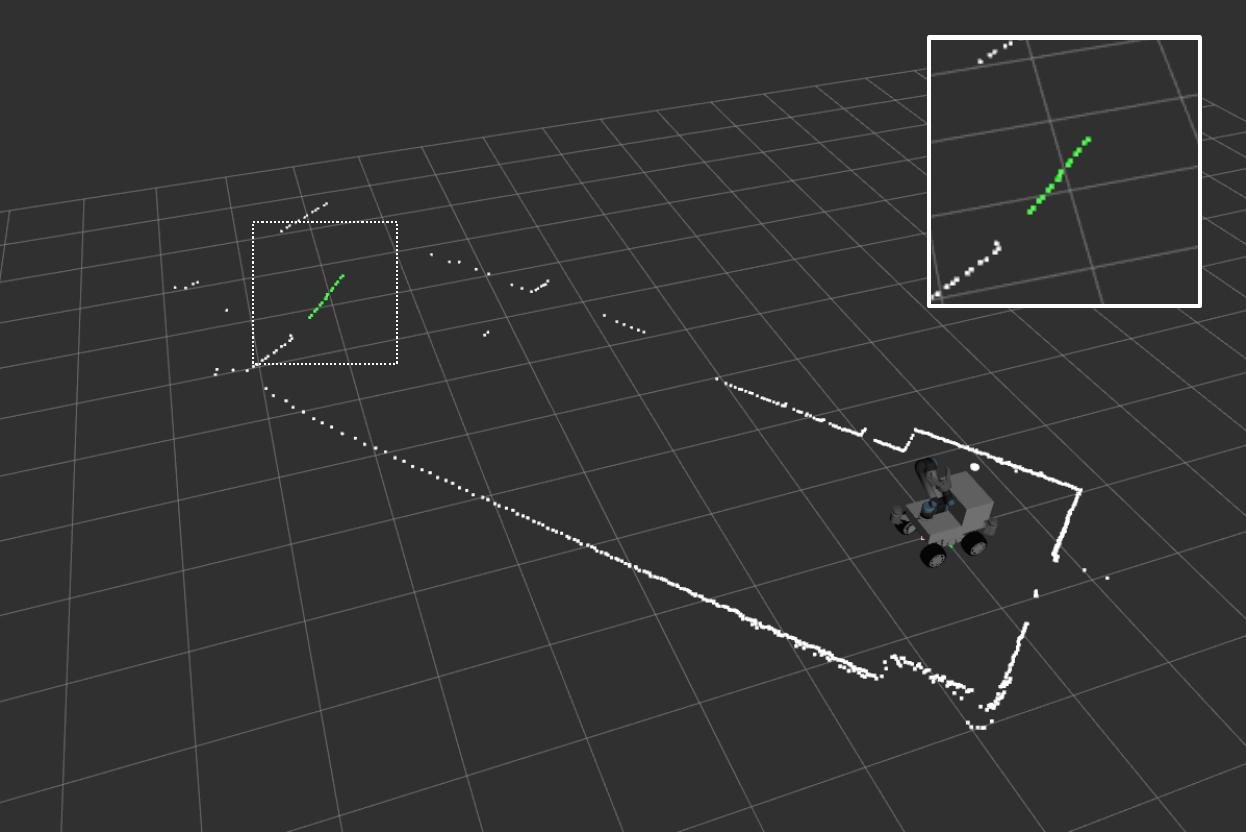}
  }
  \caption{Panel detection. (a) The proof that the robot is able to correctly detect the panel from 8 meters (the panel is the line of \textit{green} laser points). (b) The same detection from 5 meters.}
  \label{fig:panel_detection}
\end{figure}

Laboratory experiments (see Figure~\ref{fig:panel_detection}) show that the robot is able to correctly identify the pose of the panel until a distance of 8~m and with an orientation up to $\pm 25\degree$ with respect to the robot.

\begin{figure}[t]
\centering
  \subfloat[]{
    \includegraphics[width=0.43\textwidth]{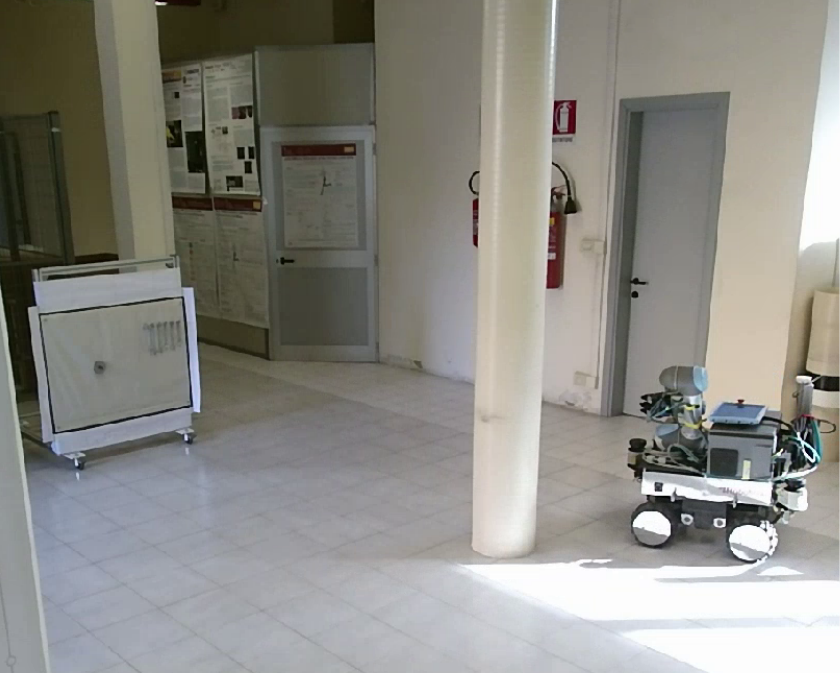}
  }\quad
  \subfloat[]{
    \includegraphics[width=0.43\textwidth]{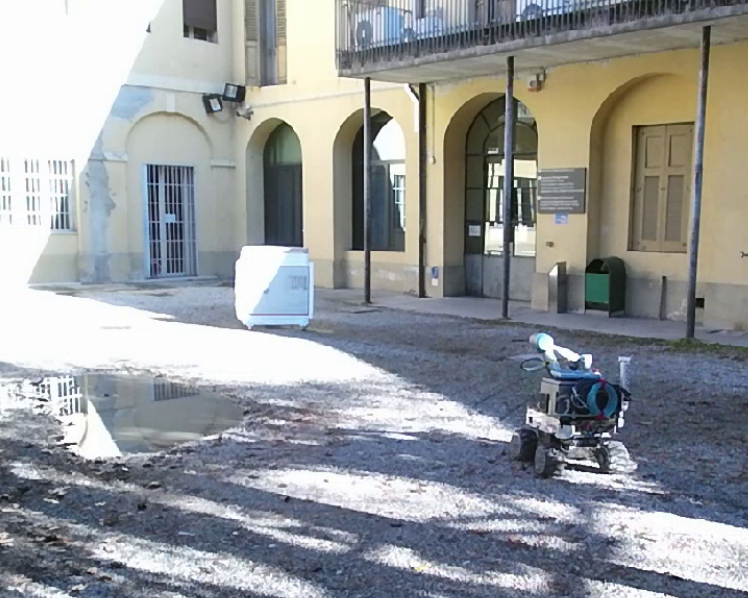}
  }
  \caption{RUR53 while it is approaching the panel. (a) Indoor. (b) Outdoor.}
  \label{fig:docking}
\end{figure}

\begin{figure}
\centering
\includegraphics[width=.43\textwidth]{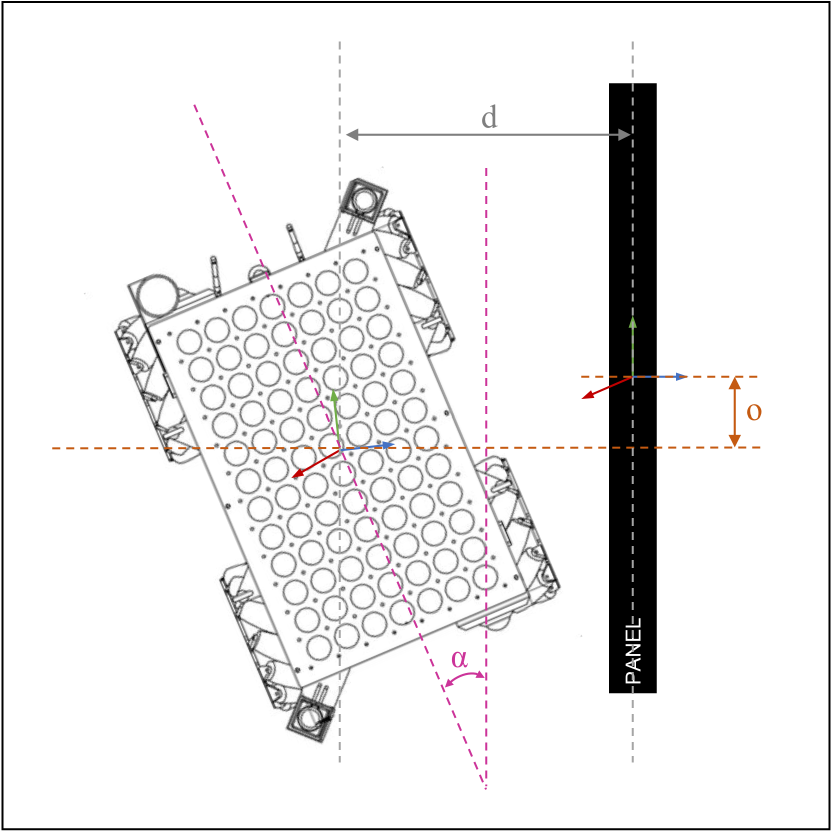}
\caption{Docking parameters. $d$ is the distance (in meters) between the robot and the panel expressed along the \textit{z} axis (\textit{blue} arrow); $o$ is the offset (in meters) separating the two reference systems along the \textit{x} axis (\textit{green} arrow); $\alpha$ is the docking angle: the angle between the robot and the panel (in degrees).}\label{fig:robot_orientation}
\end{figure}

Once the panel is identified and its pose is estimated, the robot has to approach it as precisely as possible. We evaluated the \textit{z}-distance $d$ separating the mobile base from the panel (in meters); the \textit{x}-offset $o$ separating the two reference systems (in meters); and the docking angle $\alpha$ (in degrees) (see Figure \ref{fig:robot_orientation}).
Table~\ref{tbl:docking_performances} shows the results obtained in 10 lab trials. The lab floor is smooth and tiled. An average error of 0.02\,m can be noticed with respect to $d$, confirming that the proximity constraint is almost always achieved. With respect to $o$, an average error of 0.15\,cm has been computed, with a maximum error of 0.2\,m. This value is high and, most of the time, it is negative because of the high load that the robot has to transport. Focusing on $\alpha$, an average error of 9\degree ~is observed, with a maximum error of 15\degree ~due to the friction of the wheels, which have to carry a high load. Tests were performed also outside the laboratory (see Figure \ref{fig:docking}).

\subsection{Wrench and Valve Classifiers}

\subsubsection{Wrench and Valve Datasets}

Wrench and valve datasets are acquired by means of two cameras, i.e., a Bumblebee2 and a Grasshopper3. Images are taken under different light conditions, indoors and outdoors, and at different hours of the day, to include multiple reflection conditions and shadows. The Hard Negative Mining~\cite{liao2007learning} technique is used to repeatedly add false positive and false negative samples, thus increasing the datasets and improving the performance of the classifier.

\paragraph{Wrench Dataset}

The wrench dataset is composed of 830 positive samples and 7350 negative samples. While acquiring the images, four different sets of wrenches are exploited to avoid the overfitting of a specific wrench model. Positive samples are represented by 100$\times$100~pixel images. This format guarantees a good balance between high resolution and reduced training time. Negative samples, instead, depict backgrounds and other objects, such as the valve, the shadows that wrenches form on the panel, and those metallic supports to which wrenches are attached.

\paragraph{Valve Dataset} 
Images of different orientations of the valve and different rotations of the stem form the valve dataset. A set of 450 $100\times100$~pixel valve images represents the collection of positive samples. 7490 background images, instead, define the negative samples.

\subsubsection{Performance Measurement}

\begin{table}
  \tbl{Docking performances. See Figure~\ref{fig:robot_orientation} for the parameter legend.}
  {\begin{tabular}{lcccccc}
        \toprule
        &Desired Value&Average&Median&Max&Min&Std.Dev.\\
        \midrule
        d (m)&0.800&0.786&0.785&0.814&0.765&0.017\\ 
        o (m)&0.000&-0.147&-0.174&-0.082&-0.202&0.060\\ 
        $\alpha$ (deg)&0.00&9.38&7.80&15.53&5.41&4.22\\ 
        \bottomrule
    \end{tabular}}
    \label{tbl:docking_performances}
\end{table}

In order to evaluate the performance of the trained classifiers, the following elements are analyzed:
\begin{equation}
\label{eq:accuracy}
\mathit{accuracy} = \frac{\mathit{TP} + \mathit{TN}}{\mathit{TP} + \mathit{TN} + \mathit{FP} + \mathit{FN}}\,,
\end{equation}
\begin{equation}
\label{eq:precision}
\mathit{precision} = \frac{\mathit{TP}}{\mathit{TP} + \mathit{FP}}\,,
\end{equation}
\begin{equation}
\label{eq:recall}
\mathit{recall} = \frac{\mathit{TP}}{\mathit{TP} + \mathit{FN}}\,,
\end{equation}
\begin{equation}
\label{eq:F2}
\mathcal{F}_2 = \frac{(1 + 2^2) * \mathit{precision} * \mathit{recall}}{2^2 * \mathit{precision} + \mathit{recall}}\,.
\end{equation}
where $\mathit{TP}$, $\mathit{TN}$, $\mathit{FP}$, $\mathit{FN}$ are the true positive, true negative, false positive, and false negative counts. The $\mathcal{F}_2$ score~\cite{Rijsbergen:1979:IR:539927} is an harmonic average of \textit{precision} and \textit{recall} which promotes high \textit{recall}, i.e., a low number of false negatives.

The evaluation has been performed on both specific test sets and data collected during the Challenge. They include 25 test sets for the wrench images and 25 test sets for the valve images, obtained by randomly splitting the corresponding dataset into a 70\%  used for training and a 30\% used for test.
As depicted in Table~\ref{tbl:classifiers_performance}, tests performed on these test sets show that the wrench classifier is characterized by a high precision (97.0\%) and an high recall (93.6\%), which let achieve a $\mathcal{F}_2$ score of 94.3\%. The valve classifier, instead, is characterized by a precision of 98.2\% and a recall of 94.1\%, reflecting a $\mathcal{F}_2$ score of 94.9\%. 
Table~\ref{tbl:classifiers_challenge} shows the performance of the classifiers when applied to the 255 wrench samples and the 40 valve samples collected during the Challenge. In this case, the wrench classifier has a true positive rate of 100\% and a false positive rate of 0\%. With respect to the valve classifier, a true positive rate of 90\% and a false positive rate of 10\% are observed.

\subsection{Wrench Recognition and Grasping}

The wrench recognition and grasping routines are evaluated in the lab by asking the robot to correctly recognize and pick up a target wrench for 50 trials. In order to acquire representative data, sets of 6 wrenches of 10 types are hung to the panel. At each trial, their pose changes as well as the target wrench.

According to the Challenge specifications, the maximum score is assigned if the robot correctly recognizes and grasps the tool (\textit{Correct Grasp}). Different scores are assigned if the wrench is correctly recognized (\textit{Correct Recognition}) or correctly recognized and grasped, but not so accurately to operate on the valve (\textit{Grasp}). In the worst case, the robot loses the tool while grasping it (\textit{Loss}). 
Table~\ref{tbl:wrench_manipulation} summarizes the performance obtained in the laboratory: the proposed system correctly recognizes the wrench 92\% of the time and it correctly grasps it 70\% of the time. 
16\% of time, instead, the robot picks up the wrench but the tool is not correctly grasped, making it impossible to operate on the valve.
Thus, the system scores points 86\% of time.

Table~\ref{tbl:wrench_recognition_time} summarizes the execution time obtained by the wrench recognition algorithm during the 50 reps.
Input images are black and white and have a resolution of 964$\times$724~pixels. 
On average, RUR53 finds the right wrench in 3.338\,s; in the worst case, it needs 3.770\,s.

Experiments prove the reliability and good performance of the wrench recognition and grasping routines.
Focusing on the wrench detector, the main reason of failure is the outdoor light condition: a very bright light causes strong reflections that can lead to a lacking 3D reconstruction of the working area or to a wrong wrench detection.
A lacking 3D reconstruction can cause both a wrong measure of the wrench head, that consequently cause an erroneous recognition of the searched wrench, and a false estimation of the grasping point.
On the grasping side, 
errors in estimating the grasp point or oscillations during the grasp may cause the gripper to loose the wrench. These oscillations are strong if tests are performed outside and, for example, there is wind.

\begin{table}[t]
    \tbl{Performance of wrench and valve classifiers on their test sets.}
    {\begin{tabular}{lcccc}
        \toprule
        Classifier&Accuracy&Precision&Recall&$\mathcal{F}_2$\\
        \midrule
        Wrench&94.1\%&97.0\%&93.6\%&94.3\%\\
        Valve&95.1\%&98.2\%&94.1\%&94.9\%\\
        \bottomrule
    \end{tabular}}
    \label{tbl:classifiers_performance}
\end{table}

\begin{table}[t]
    \tbl{Performance of wrench and valve classifiers obtained during the challenge.}{
    \begin{tabular}{cccc}
        \toprule
        Classifier&True Positive&False Positive&Samples\\
        \midrule
        Wrench&100\%&0\%&225\\
        Valve&90\%&10\%&40\\
        \bottomrule
    \end{tabular}}
    \label{tbl:classifiers_challenge}
\end{table}

\begin{table}[t]
    \tbl{Performance of the wrench detection and grasping routines on 50 reps.}
    {\begin{tabular}{cccc}
        \toprule
        Correct Recognition&Correct Grasp&Grasp&Loss\\
        \midrule
        92\%&70\%&16\%&14\%\\
        \bottomrule
    \end{tabular}}
    \label{tbl:wrench_manipulation}
\end{table}

\begin{table}[t]
    \tbl{Execution time of the wrenches recognition algorithm. Measures are in seconds.}
    {\begin{tabular}{lccccc}
        \toprule
        &Average&Median&Maximum&Minimum&Std. Dev.\\
        \midrule
        Wrench detection (s)&3.338&3.305&3.770&3.266&0.106\\
        \bottomrule
    \end{tabular}}
    \label{tbl:wrench_recognition_time}
\end{table}

\subsection{Valve Detection and Manipulation}

The robot has to insert the wrench into the valve stem and rotate it clockwise 360\degree. To do that, the orientation of the stem should be estimated.

In the lab, 50 trials have been performed making the robot estimate four stem angles $\alpha$: 0\degree, 15\degree, 30\degree ~and 45\degree. 
Table~\ref{tbl:valve_manipulation} shows obtained results in terms of average error $e = \frac{1}{N}\sum\limits_{i=1}^N (x_i - \alpha_i)$, where $N=4$ is the number of target angles and $\alpha_i$ is the i-th target angle. If $x_i$ represents the \textit{average angle}, than $e$ is equal to $0.6875\degree$. The \textit{maximum average error} is equal to $1.99\degree$ and the \textit{minimum average error} is $0.5125\degree$. 
Performance obtained during the rest of the task confirms that this error range guarantees the correct insertion of the wrench into the valve stem for more than half the times. In detail, the insertion routine gives successful results 53.3\% of the time. This outcome results from 30 trials, with the manipulator arm starting from a random point and the valve stem randomly rotated.
Errors are mostly due to light reflections and wrong alignments of the cameras with the valve. Moreover, the insertion of the wrench into the valve is affected by the movement of the shock absorbers during the manipulator extension.
Once the wrench is correctly engaged to the stem, the robot completely rotates the valve 86.7\% of the time. Failures come  from a misalignment of the wrench to the stem. In such a case, the wrench loses the stem engagement while rotating or the exacted effort causes the protective stop of the robotic arm. In any case, on 15 trials, a rotation of at least 90\degree \ is always performed.

Table~\ref{tbl:valve_detection_time} summarizes the execution time of the valve detection algorithm. As for the wrench recognition routine, input images are black and white and have a resolution of 964$\times$724~pixels.
On average, on 50 trials, RUR53 needs 0.059\,; 0.071\,s are required in the worst case.

\begin{table}[t]
    \tbl{Performance of the valve detection routine on 50 reps. Measures are in degrees.}
    {\begin{tabular}{lccccc}
        \toprule
        $\alpha$ (deg)&Average&Median&Maximum&Minimum&Std. Dev.\\
        \midrule
        0.00&0.23&0.00&0.73&0.00&0.44\\
        15.00&16.51&16.39&17.53&15.96&0.46\\
        30.00&30.79&30.65&33.69&29.20&1.14\\
        45.00&45.24&45.00&46.01&45.00&0.43\\
        \bottomrule
    \end{tabular}}
    \label{tbl:valve_manipulation}
\end{table}

\begin{table}
    \tbl{Execution time of the valve detection algorithm. Measures are in seconds.}
    {\begin{tabular}{lccccc}
        \toprule
        &Average&Median&Maximum&Minimum&Std. Dev.\\
        \midrule
        Valve detection (s)&0.059&0.059&0.071&0.052&0.005\\
        \bottomrule
    \end{tabular}}
    \label{tbl:valve_detection_time}
\end{table}

\section{Challenge Performances and Lessons Learned}\label{lessons_learned}
Lab and Challenge tests demonstrate that the proposed system has a good potential. 
It correctly recognizes the assigned wrench, even when different from the training ones, it correctly detects the valve stem, and it can complete the manipulation routine. However, 
reaching almost the maximum payload of the mobile base made difficult to maintain the laser scan plane parallel to the ground and, in Abu Dhabi, the arena floor was confused with the  barriers, negatively affecting the localization of the robot within the arena. Furthermore,
the panel was black and the temperature was high (more than 40\degree C); these constraints led the laser to a partial detection of the panel and pushed our team to adopt a semi-autonomous mode during the Grand Challenge. 

Obtained results show that harmonizing technical requirements with the necessity of a low-cost robot is challenging. 
Indeed, adopting a mobile base with a higher payload or a larger support polygon would have reduced the need for balance control, thus minimizing the unbalancing of the robot and having omnidirectional wheels would have allowed a faster and more precise robot motion --- unfortunately, at the time we chose the robot base for the competition we were expecting an unpaved arena, not suitable for omnidirectional wheels.
Moreover, mounting a LiDAR more suitable for dark surfaces would have helped finding the panel as well as combining data from LiDARs with data from a Grayscale or RGB camera or using a 3D LiDAR able to produce a dense point cloud of the environment, which would have improved also the wrench detection and manipulation. 
Finally, even if the implemented wrench and valve recognition routines were performing, adopting a visual servoing approach could led to a further performance improvement.

\section{Conclusion}\label{conclusion}
This paper presented RUR53, a UGV able to autonomously navigate outdoors, recognize work tools, and perform manipulation tasks.
RUR53 was tailored to face Challenge 2 of MBZIRC 2017, which asked a UGV to autonomously navigate within an arena, identify and reach a panel, recognize and manipulate a wrench located on it and use this wrench to operate a valve stem located on the panel itself. 

The proposed system provides a modular software architecture that allows the robot to fast adapt to new environments, detect objects other than those for which it has been trained, and make easy the replacements of modules to perform different tasks.
The ability to rapidly switch to a semi-autonomous mode during the challenge proves that such architecture is a winning choice to face unexpected situations/occurrences.
Lab experiments demonstrate that the robot can autonomously detect and reach the panel from a distance of 8 meters and it docks with an average orientation error of about 9\degree. 
In the lab, an accuracy of about 95\% characterized the detection and recognition of both the correct wrench and the valve stem, while a 100\% success is achieved during the Challenge.
Despite the good performance of the autonomous perception system, the payload of the mobile base and the detection capability of 2D LiDARs lead to some issues in detecting the panel. 
Thus, during the competition, a semi-teleoperated mode was implemented. This choice let us rank third in the Grand Challenge in collaboration with the Czech Technical University in Prague, the University of Pennsylvania, and the University of Lincoln (UK).

Lessons learned gave us the capability to deal with complex industrial scenarios, like the one proposed by the H2020 CleanSky2 EURECA project~\cite{roveda2017eureca,roveda2018human}. 
EURECA asks a mobile manipulator to localize some components, bring them inside a real A320 aircraft fuselage, and use them to assembly the aircraft side-wall both in an autonomous and in a collaborative mode. 

\section*{Acknowledgments}
We would like to thank the MBZIRC organizing committee, and the Khalifa University, for the financial support and for giving us the opportunity to participate in this event. Thanks to IT+Robotics Srl and EXiMotion Srl for the financial support. Thanks to SICK AG and NVIDIA Inc. for the hardware equipment they provided us. A special thanks to all the students who worked on the project: Enrica Rossi, Alex Badan, Luca Benvegn\`u, Tobia Calderan, Riccardo Fantinel, Thomas Gagliardi, Leonardo Pellegrina, Matteo Tessarotto and Marco Zaghi.



\section*{Founding}
This work was partially supported by the Khalifa University of Science, Technology and Research (Abu Dhabi, UAE) under Contract 2016-MBZIRC-13.

\bibliographystyle{tfnlm}
\bibliography{bibliography}

\end{document}